\documentclass{article}

\usepackage[utf8]{inputenc}
\usepackage{textgreek}

\usepackage{microtype}
\usepackage{graphicx}
\usepackage{subcaption}
\usepackage{booktabs} 

\usepackage{newfloat}
\DeclareFloatingEnvironment[name=Extended Data Figure, placement=h!]{extendeddata}

\usepackage{siunitx} 

\usepackage{hyperref}

\usepackage{amsmath}
\usepackage{amssymb} 

\PassOptionsToPackage{sort&compress}{natbib} 
\usepackage[accepted]{icml2026}
\setcitestyle{numbers,square,comma}

\usepackage{amsmath}
\usepackage{amssymb}
\usepackage{mathtools}
\usepackage{amsthm}

\usepackage[version=4]{mhchem}

\usepackage[capitalize,noabbrev]{cleveref}

\theoremstyle{plain}

\theoremstyle{definition}

\theoremstyle{remark}

\usepackage[textsize=tiny]{todonotes}

\icmltitlerunning{Dynamic language model representations for multi-objective reaction optimisation}

\begin{document}

\twocolumn[
  \icmltitle{Dynamic language model representations for multi-objective reaction optimisation}



  \icmlsetsymbol{equal}{*}

  \begin{icmlauthorlist}
    \icmlauthor{Joshua W. Sin}{equal,PTDC,LIAC}
    \icmlauthor{David Ming Segura}{equal,PTDC,LIAC,NCCR}
    \icmlauthor{Bojana Rankovi\'{c}}{equal,LIAC,NCCR}
    \icmlauthor{Siu Lun Chau}{EPIC}
    \icmlauthor{Marius D. R. Lutz}{pred}
    \icmlauthor{Andrea Anelli}{pred}
    \icmlauthor{Ryan P. Burwood}{SSS}
    \icmlauthor{Kurt P\"{u}ntener}{PTDC}
    \icmlauthor{Maximilian J. Notheis}{PTDC}
    \icmlauthor{Raphael Bigler}{PTDC}
    \icmlauthor{Philippe Schwaller}{LIAC,NCCR}
  \end{icmlauthorlist}

  \icmlaffiliation{PTDC}{Process Chemistry \& Catalysis, Synthetic Molecules Technical Development, F. Hoffmann-La Roche AG, Basel, Switzerland}
  \icmlaffiliation{LIAC}{Laboratory of Artificial Chemical Intelligence (LIAC), EPFL, Lausanne, Switzerland}
  \icmlaffiliation{NCCR}{National Centre of Competence in Research (NCCR) Catalysis, EPFL, Lausanne, Switzerland}
  \icmlaffiliation{EPIC}{Epistemic Intelligence \& Computation Lab, College of Computing \& Data Science, Nanyang Technological University, Singapore}
  \icmlaffiliation{SSS}{Solid State Sciences, Synthetic Molecules Technical Development, F. Hoffmann-La Roche AG, Basel, Switzerland}
  \icmlaffiliation{pred}{Roche Pharma Research and Early Development (pRED), F. Hoffmann-La Roche AG, Basel, Switzerland}

  \icmlcorrespondingauthor{Joshua W. Sin}{wing\_pong.sin@roche.com}
  \icmlcorrespondingauthor{David Ming Segura}{david.segura@epfl.ch}
  \icmlcorrespondingauthor{Bojana Rankovi\'{c}}{bojana.rankovic@epfl.ch}
  \icmlcorrespondingauthor{Philippe Schwaller}{philippe.schwaller@epfl.ch}

  \icmlkeywords{Machine Learning, ICML, Bayesian optimization, Large language models, Process chemistry, Chemical reaction optimization}

  \vskip 0.3in
]



%
%

\printAffiliationsAndNotice{\icmlEqualContribution}

\begin{abstract}

Optimising chemical reactions across multiple objectives, such as yield, selectivity, and safety, is central to chemical synthesis, and model-driven approaches depend critically on how reaction components are represented. Established featurisations are either chemically uninformative, as with one-hot encodings, or, as with molecular descriptors, do not readily extend across chemically distinct components. For structurally and functionally diverse components, it is therefore unclear what a shared representation should contain. Constructing such a representation is itself a challenging research undertaking that must be revisited for each new reaction system. Here we bypass this step by learning the reaction representation dynamically from text. Textual descriptions of reaction conditions are encoded by a fine-tuned language model trained jointly with Gaussian process surrogates, yielding task-adaptive representations within a multi-objective Bayesian optimisation loop. Across nickel- and palladium-catalysed cross-couplings in both sequential and parallel experimentation regimes, this approach reaches optimisation convergence in fewer experiments than descriptor libraries or one-hot encoding. Applied prospectively to a palladium-catalysed cyanation spanning mixed ligand denticity and heterogeneous additives, and to a three-objective asymmetric hydrogenation across chiral iridium and ruthenium catalyst families, two rounds of high-throughput experimentation (192 reactions, under 3\% of each design space) delivered conditions translating directly to gram scale in 94\% and 84\% isolated yield, the latter at 99.6\% enantiomeric excess.

\end{abstract}

\section{Introduction}

Chemical reaction optimisation is essential to chemical synthesis, and optimising reaction conditions requires navigating vast combinatorial spaces of reaction components such as ligands, catalysts, and bases, while balancing multiple objectives such as yield, selectivity, and cost~\cite{taylorBriefIntroductionChemical2023, DkerBayerBO2026, BallMLReactionPrediction2025, NippaCompoundSynthesisDigitalisation2025, SchoepferCostBO2024}. This challenge is particularly acute in pharmaceutical process and preclinical chemistry, where identifying robust conditions for active pharmaceutical ingredient (API) synthesis entails additional environmental, health, and safety considerations~\cite{zhangProcessChemistryScience2006, LiptonProcessChem2006, ZhaoNonPGM2024, ClarkeSustainable2018}, and in academic settings where enabling novel transformations often requires extensive exploration of many diverse reaction parameters~\cite{RowsellFeCatalysis2024, DelaneyScienceCu2023}.

Machine learning approaches have proved remarkably effective, with Bayesian optimisation emerging as the leading framework for data-efficient reaction optimisation. It has been applied in low-data regimes~\cite{shieldsBayesianReactionOptimization2021, torresMultiObjectiveActiveLearning2022, Braconi2023, RomerNickelAlkeneEDBO2024, CadgeNickelDielsEDBO2025, ZhangSchottenqNEHVI2024, TaylorMultiTask2023, DunlapContFlowBO2023} and, more recently, in parallel high-throughput settings~\cite{sinHighlyParallelOptimisation2025, AlphaPSO2026}. Despite these algorithmic advances, how best to represent reactions for model-driven optimisation remains an open and consequential challenge.

\begin{figure*}[t!]
    \centering
    \includegraphics[width=1\linewidth]{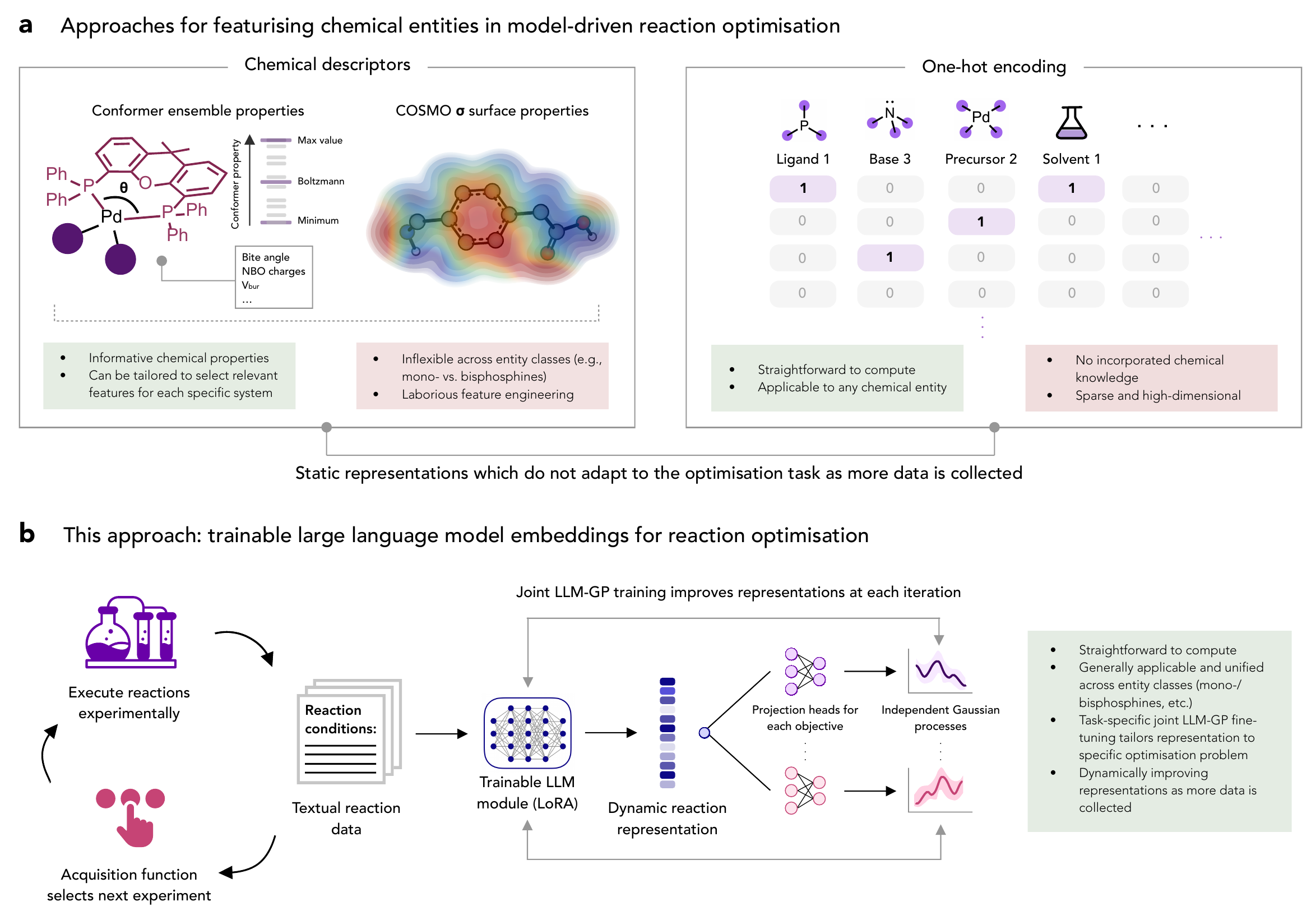}
    \caption{\textbf{Featurisation approaches for model-driven reaction optimisation.} \textbf{a,} Conventional approaches for representing chemical entities. Left: molecular descriptors (e.g., Sterimol parameters, buried volume, and bite angle) incorporate physicochemical properties derived from density functional theory (DFT)~\cite{KohnDFT1996} or extended tight-binding (xTB)~\cite{BannwarthXTB2019}, enabling chemical similarity-aware representations. However, descriptor sets are inflexible across entity classes, require domain-specific expertise for feature engineering, and can be computationally expensive to generate. Right: one-hot encoding assigns a binary indicator vector to each categorical component (e.g., different ligands, bases, and precursors), yielding sparse, high-dimensional representations that encode no chemical similarity. \textbf{b,} This approach: trainable large language model (LLM) embeddings for reaction optimisation. Textual descriptions of reaction conditions are encoded by a LoRA-fine-tuned~\cite{LoRA2021} language model into dense, continuous embeddings. These embeddings are passed through objective-specific projection heads into independent Gaussian process (GP) surrogate models, and the entire architecture (LLM-GP) is trained jointly. The architecture is embedded within a multi-objective Bayesian optimisation cycle, with the reaction representations improving at each iteration, organising representations by reaction performance. This yields a unified, task-adaptive representation without requiring descriptor computation or domain-specific feature engineering.}
    \label{fig:figure_1}
\end{figure*}

Molecular descriptors have been the dominant featurisation strategy in reaction optimisation and prediction (\Cref{fig:figure_1}a). Pre-computed descriptor libraries such as Kraken~\cite{genschComprehensiveDiscoveryPlatform2022} and the COSMO-RS~\cite{moityPanoramaSustainableSolvents2012} database provide readily available steric and electronic properties for established compound classes and have been widely adopted~\cite{daltonUtopiaPointBayesian2024, sinHighlyParallelOptimisation2025, AlphaPSO2026, RomerNickelAlkeneEDBO2024, CadgeNickelDielsEDBO2025, CadgeBisphosphine2025, OcampoMolli2024, ChristensenDataAutonomous2021}. Alternatively, tailored and often laborious feature engineering workflows, involving computationally expensive density functional theory (DFT) or semi-empirical calculations, conformer ensemble analysis, and reaction-specific mechanistic interrogation, can yield informative descriptors~\cite{SamhaUllmann2024, DotsonBisphosphine2022, SouzaPhotoRedoxDFT2025, GandhiChanLam2025, RinehartDPT2023, CallEnantio2025, Ocampo2025, BannwarthXTB2019, KohnDFT1996}.

Despite their widespread use, descriptor-based approaches face fundamental limitations. Selecting which descriptors to compute relies on a priori chemical intuition about which molecular properties govern reactivity, a process that is biased by existing mechanistic understanding and must often be revisited for each new reaction system. More critically, descriptors are often not transferable across chemically distinct reaction components. Steric and electronic parameters for monodentate phosphines, for example, are only partially applicable to bidentate ligands, and reactions involving chemically heterogeneous components such as inorganic bases and organometallic additives may lack any meaningful shared descriptor space, making feature engineering increasingly prohibitive as chemical diversity grows.
One-hot encoding offers a simpler alternative that is universally applicable and has shown competitive performance in some settings~\cite{RankoviOHE2024, shieldsBayesianReactionOptimization2021, torresMultiObjectiveActiveLearning2022}. However, it produces sparse, high-dimensional representations that encode no chemical similarity, and its dimensionality scales unfavourably with the number of reaction components~\cite{sinHighlyParallelOptimisation2025, RankoviOHE2024} (\Cref{fig:figure_1}a).

Advances in large language models (LLMs) have shown strong performance across chemical tasks~\cite{AshyrmamatovLLMSurvey2026,OargaLLMOntology2026, MBranChemCrow2024, ZimmermannLLMExamples2025, LuT5Chem2022, AlamparaGeneralChemLLMReview2026, BranSteer2026, SynthStrategy, WieczorekTransferLearning2025} and demonstrated that their learned embeddings can serve as dense, informative representations for predictive modelling~\cite{RossMolFormer2022, ChemBerta, JablonkaLLMPredChem2024, SeguraCheMatE}. Recent work has further shown that LLM representations can be adapted to specific optimisation tasks through joint fine-tuning with Gaussian process surrogates, demonstrated for single-objective, sequential optimisation~\cite{GoLLUMUpdated}. Crucially, LLMs can encode arbitrarily complex chemical information into fixed-length representations directly from text, offering a naturally unified featurisation method across component classes. These important advances notwithstanding, a general framework for multi-objective reaction optimisation, applicable across chemically heterogeneous reaction components and validated across experimental regimes from low-data to high-throughput settings, has yet to be realised.

Here, we introduce Alice, a multi-objective chemical reaction optimisation framework that uses trainable LLM embeddings as a unified reaction representation (\Cref{fig:figure_1}b). Textual descriptions of reaction conditions are encoded by a low-rank adaptation (LoRA)-fine-tuned language model and passed through reaction objective-specific projection heads into independent Gaussian process surrogates, with the architecture trained end-to-end via joint marginal log-likelihood maximisation. This jointly trained architecture, hereafter referred to as the LLM-GP framework, produces dynamic embeddings that are refined at each optimisation iteration, adapting the reaction representation to the optimisation task at hand. We first benchmark the LLM-GP framework retrospectively against established descriptor libraries and one-hot encoding baselines across sequential low-data optimisation of nickel- and palladium-catalysed cross-couplings and large-batch 96-well plate high-throughput experimentation campaigns, converging on high-performing conditions in fewer experiments across all settings. 

We then apply the framework prospectively to two wet-lab campaigns executed on an automated high-throughput platform, each chosen for a form of chemical heterogeneity for which descriptor libraries are not readily available. In a palladium-catalysed cyanation, a single representation spans mono- and bidentate phosphine catalysts alongside additives ranging from elemental zinc to organic and inorganic bases; within this space the framework identified conditions using a lower-hazard cyanide source that gave $>$99\% conversion and $>$99\% selectivity, and 94\% isolated yield on gram scale. In an asymmetric dynamic kinetic hydrogenation, the framework balanced conversion, diastereomeric excess (de), and enantiomeric excess (ee) across chiral catalysts spanning different metals, donor sets, and coordination geometries, delivering the target \textit{syn} alcohol in 84\% isolated yield and 99.6\% ee on scale-up. Both campaigns used only two rounds of 96 experiments, sampling under 3\% of their respective design spaces. Neither required descriptor computation or feature engineering, indicating that effective multi-objective optimisation is achievable in reaction spaces where constructing a shared descriptor representation would itself be a substantial undertaking.

\begin{figure*}[h!]
    \centering
    \includegraphics[width=1\linewidth]{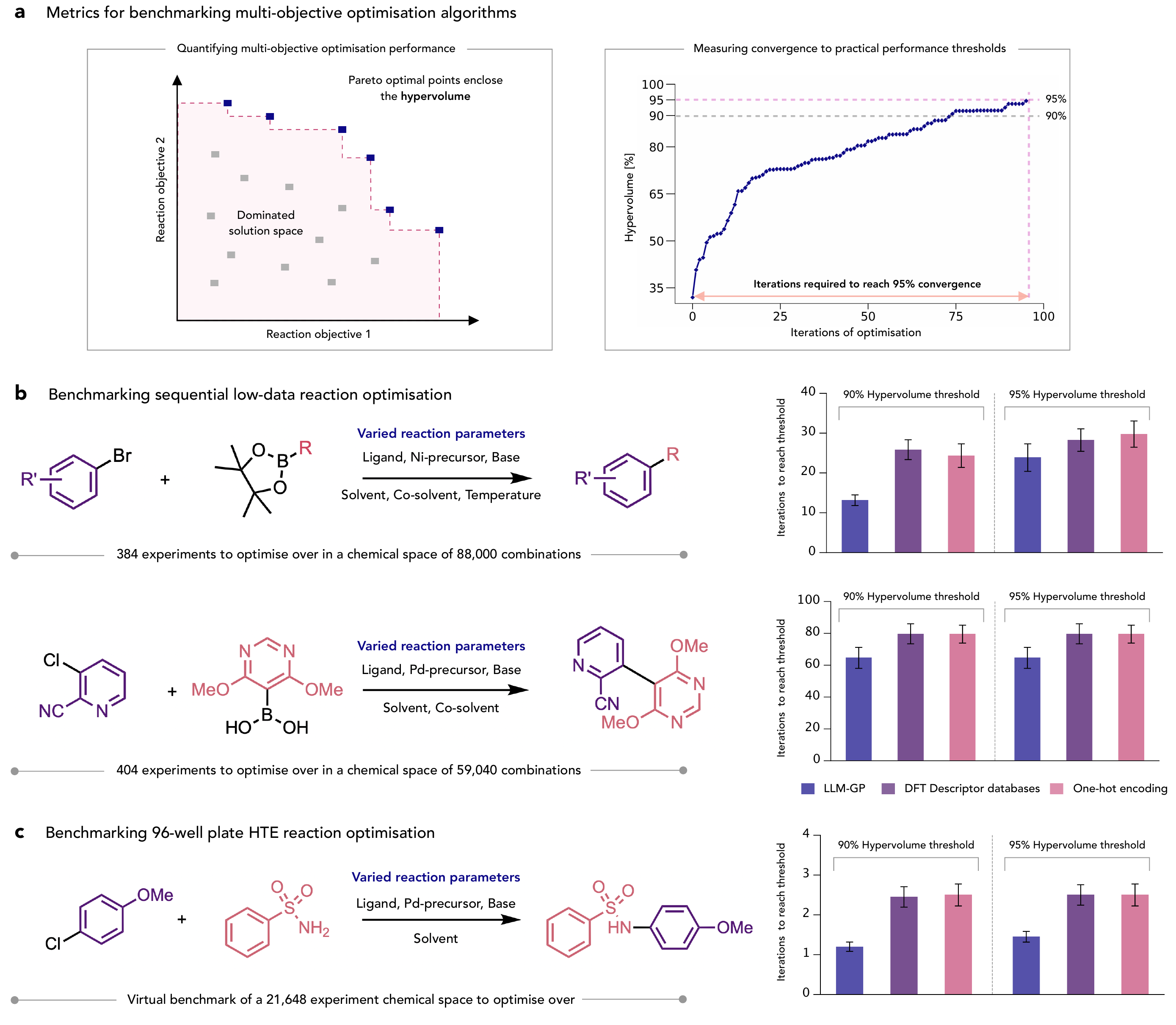}
    \caption{\textbf{Benchmarking multi-objective reaction optimisation across experimental regimes.} \textbf{a,} Metrics for evaluating multi-objective optimisation. Left: the hypervolume indicator quantifies the volume of objective space dominated by the current Pareto-optimal set, providing a scalar measure of multi-objective performance. Right: convergence is assessed by tracking the normalised hypervolume over optimisation iterations; the number of iterations required to reach practical performance thresholds (90\% and 95\% of maximum hypervolume) serves as a comparative metric across featurisation methods. The percentage of optimisation runs (initialised with different random seeds) reaching each threshold is reported in Extended Data \Cref{SI_sec:main_text_supp}. 
    \textbf{b,} Sequential low-data reaction optimisation benchmarks. Two multi-objective case studies from published reaction datasets are evaluated: a nickel-catalysed Suzuki coupling~\cite{sinHighlyParallelOptimisation2025} and a palladium-catalysed Suzuki coupling~\cite{AlphaPSO2026}. Iterations required to reach 90\% and 95\% hypervolume thresholds are compared for the LLM-GP framework, DFT descriptor libraries (Kraken~\cite{genschComprehensiveDiscoveryPlatform2022}, COSMO-RS~\cite{moityPanoramaSustainableSolvents2012}), and one-hot encoding, repeated across 20 random seeds (see Methods for more details). Runs not reaching a threshold within the experimental budget were assigned the maximum budget. 
    \textbf{c,} 96-well plate HTE reaction optimisation benchmark. A palladium-catalysed sulfonamide coupling~\cite{AlphaPSO2026} (virtual benchmark of 21,648 experiments) from a published dataset is optimised in large parallel batches of 96 experiments.}
    \label{fig:figure_2}
\end{figure*}

\section{Computational results}

We benchmarked the LLM-GP framework against established and readily available featurisation methods in reaction optimisation: DFT descriptors from the Kraken~\cite{genschComprehensiveDiscoveryPlatform2022} and COSMO RS~\cite{moityPanoramaSustainableSolvents2012} descriptor libraries, and one-hot encoding baselines. Descriptor coverage is uneven across component classes: Kraken supplies descriptors for the monophosphine ligands and COSMO-RS for the solvents, but no comparable library exists for the bases and precursors, which were therefore represented by one-hot encoding within the descriptor baseline. All three representations were embedded in the same multi-objective Bayesian
optimisation loop, using the same acquisition function (qLogNParEGO) and an identical experimental budget (see Methods).

A set of reaction conditions is Pareto optimal when no other condition improves one objective without degrading another; these points define the Pareto front (\Cref{fig:figure_2}a). Optimisation performance was measured by hypervolume, the volume of objective space enclosed by the
Pareto-optimal conditions identified so far. A larger hypervolume means the conditions found are both closer to the best achievable trade-offs and better spread across them, and we report it as a percentage of the value attained by the true Pareto front of each dataset. We compared methods by the mean number of iterations required to reach 90\% and 95\% of the maximum hypervolume, with standard error across 20 random seeds, as these thresholds represent practically relevant levels of convergence. For each dataset the optimisation budget was set by the number of iterations
needed for at least one method to reach 95\% of the maximum hypervolume on average. We additionally report the percentage of optimisation runs reaching each hypervolume threshold within the allocated experimental budget as a measure of robustness; these results are provided in Extended Data \Cref{SI_sec:main_text_supp}. Among the pre-trained language models and pooling strategies evaluated, T5-base~\cite{2020t5-basecitation} with mean pooling consistently yielded the best optimisation performance and was used throughout (full ablation studies across model architectures and pooling strategies are provided in Extended Data \Cref{sec:models-and-pooling}).

\subsection{Sequential low-data reaction optimisation}

We first evaluated our approach in sequential low-data optimisation regimes, where the primary goal is sample efficiency: minimising the number of experiments required to identify high-performing conditions (\Cref{fig:figure_2}b). We used two open-source multi-objective reaction datasets, each comprising experimental yield and selectivity as the optimisation objectives and presenting large combinatorial search spaces with diverse categorical reaction components: a nickel-catalysed Suzuki coupling (384 experiments from a space of 88,000 combinations, varying ligand, Ni-precursor, base, solvent, co-solvent, and temperature)~\cite{sinHighlyParallelOptimisation2025}, and a palladium-catalysed Suzuki coupling (404 experiments from 59,040 combinations, varying ligand, Pd-precursor, base, solvent, and co-solvent)~\cite{AlphaPSO2026}. Each campaign was initialised with 5 experiments selected using Sobol sampling to ensure diverse coverage of initial points across the search space (see Methods), after which the Bayesian optimisation loop selected one candidate per iteration.  Iteration counts below refer to these optimisation iterations and exclude the initial experiments. On the nickel-catalysed Suzuki coupling, the LLM-GP framework reached the 90\% hypervolume threshold in approximately 13 optimisation iterations on average, roughly half the number required by both the DFT descriptor-based ($\sim$26 iterations) and one-hot encoding ($\sim$24 iterations) baselines (Figure 2b). At the more stringent 95\% threshold, the gap narrowed, though the framework retained a consistent advantage, converging in approximately 24 iterations compared to $\sim$28 for DFT descriptors and $\sim$30 for one-hot encoding. On the palladium-catalysed Suzuki coupling benchmark, the framework achieved both the 90\% and 95\% thresholds in approximately 65 iterations on average, while the descriptor-based and one-hot encoding baselines required approximately 80 iterations each to reach the same performance levels (Figure 2b). Across both case studies, the LLM-GP framework consistently reached practical convergence thresholds in fewer experiments than either baseline, improving sample efficiency. Over 20 independent runs initialised with different random seeds, the framework also achieved a higher proportion of successful optimisations reaching both convergence thresholds within the allocated experimental budget, indicating more robust convergence (see Extended Data \Cref{SI_sec:main_text_supp}).

\subsection{Highly parallel HTE reaction optimisation}

Modern high-throughput experimentation (HTE) platforms have combined parallel screening with data-driven optimisation~\cite{sinHighlyParallelOptimisation2025, PorteFrugalSampling2026}, enabling the exploration of large reaction spaces in compressed experimental timescales. We next sought to evaluate our approach in this batched optimisation setting, where reducing the number of HTE plate iterations directly translates to savings in experimental time. We used an open-source palladium-catalysed sulfonamide coupling dataset (a Buchwald–Hartwig-type C–N coupling with 21,648 virtual experiments, varying ligand, Pd-precursor, base, and solvent)~\cite{AlphaPSO2026} as a benchmark, optimising yield and selectivity in parallel batches of 96 experiments to simulate 96-well plate HTE campaigns (\Cref{fig:figure_2}c). Each campaign was initialised with a Sobol-sampled plate of 96 experiments, after which each iteration selected 96 conditions at once, so one optimisation iteration corresponds to one physical plate. The LLM-GP framework reached practical convergence thresholds in approximately 1.2 plate optimisation iterations beyond initialisation, whereas both the DFT descriptor-based and one-hot encoding baselines required approximately double ($\sim$2.4) to achieve the same thresholds (\Cref{fig:figure_2}c). The framework also achieved a higher proportion of successful optimisation runs reaching both convergence thresholds, and results for 24- and 48-well plate batch sizes show the same trend (Extended Data \Cref{SI_sec:main_text_supp}). Given that each 96-well plate campaign typically requires approximately one week of experimental and analytical time, this reduction translates directly into meaningful savings in time and resources. These results demonstrate that our approach offers consistent advantages across both sequential low-data regimes, where sample efficiency is crucial, and parallel large-batch settings increasingly adopted in pharmaceutical process development and academic high-throughput screening~\cite{GotzManualHTE2023, GotzThreeComponent2025, MahjourPhactor2023}.

\begin{figure*}[h!]
    \centering
    \includegraphics[width=1\linewidth]{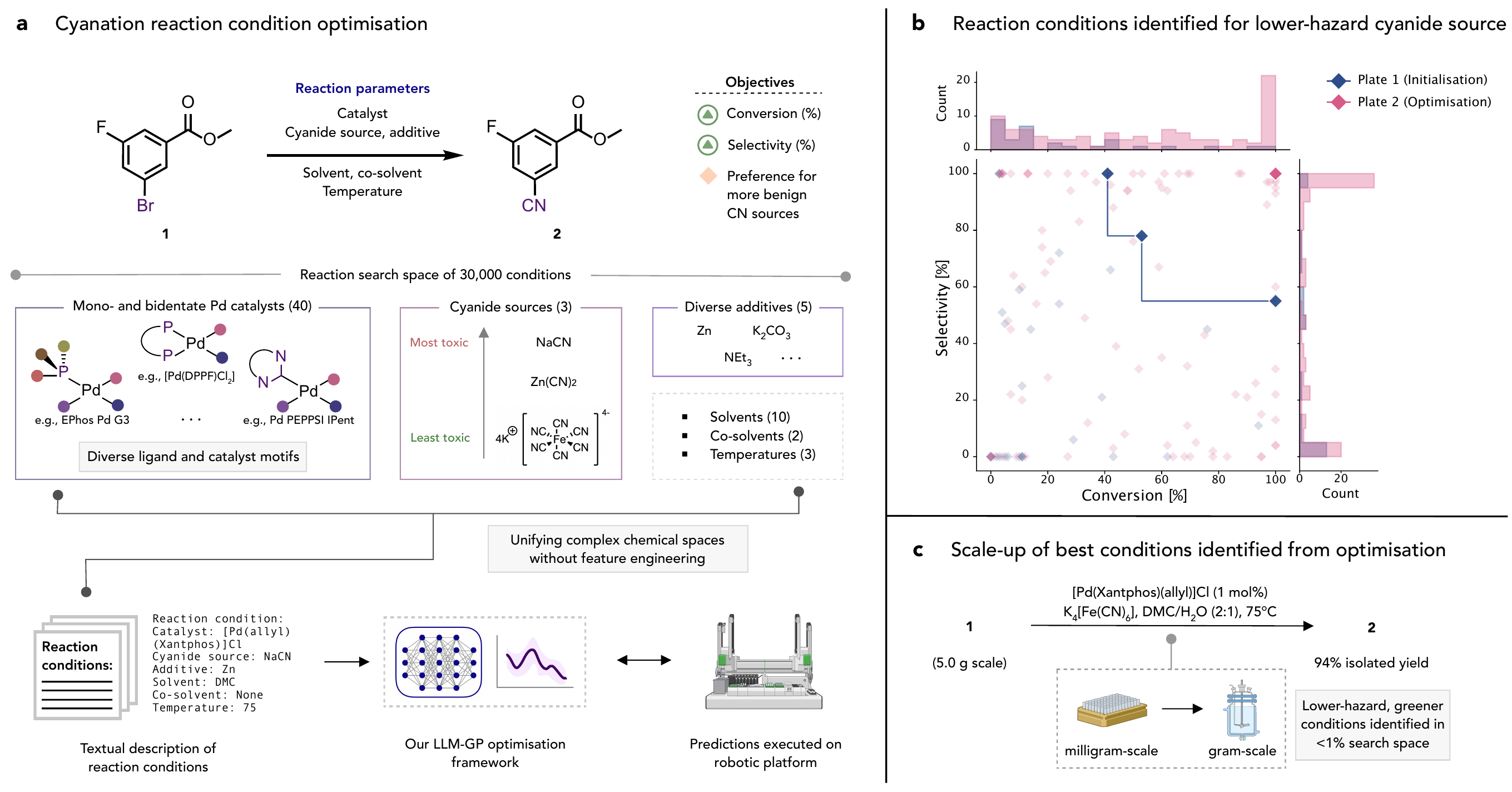}
    \caption{\textbf{Case study 1: Palladium-catalysed cyanation.}
\textbf{a,} Search space for the cyanation of \textbf{1} to \textbf{2}. Excluding reaction conditions with temperatures above the solvent boiling point results in 30,000 conditions (full search space in Extended Data Figure \ref{fig:cyanation_search_space}). Conversion (\%) and selectivity (\%) were maximised. Each reaction condition was encoded as a textual description, embedded by the joint LLM-GP framework, and model selected conditions were executed on an automated high-throughput experimentation (HTE) platform in 96-well HTE plates (see Supplementary Information Section 2.1).
\textbf{b,} 
Reaction objective distributions for HTE optimisation rounds, Plate 1 (Initialisation) and Plate 2 (Optimisation), with marginal histograms. Only conditions using \ce{K4[Fe(CN)6]}, the least toxic cyanide source, are shown on the plot. The step line traces the Pareto front among these conditions after Plate 1.
\textbf{c,} Scale-up of the best conditions identified by optimisation,
delivering \textbf{2} in 94\% isolated yield at gram scale (see Supplementary Information Section 2.2).
}
    \label{fig:pd_cyanation}
\end{figure*}

\section{Prospective experimental case studies}

Building on our retrospective benchmarks across conventional Buchwald–Hartwig and Suzuki–Miyaura datasets, we next sought to apply our framework prospectively to wet-lab reaction optimisation campaigns, moving beyond standard reaction spaces. Here, we targeted transformations containing chemically heterogeneous components that traditional reaction representations struggle to capture. In each campaign, we used our approach to identify high-performing conditions within the search space, iteratively generating ML predictions and conducting the suggested experiments with our HTE robotic platform (see Supplementary Information Section 1 for more details). All experimental data and characterisation of any isolated products are reported in the Supplementary Information.

\subsection{Case study 1: Palladium-catalysed cyanation}

Aryl nitriles are prevalent across pharmaceuticals, agrochemicals, and functional materials~\cite{NeethaNitrileReview2020, CohenMildCyanation2015}, serving as compact, metabolically stable hydrogen-bond acceptors and hydroxyl and carboxyl isosteres~\cite{FlemingNitriles2010}. Among synthetic routes accessing these motifs, the transition-metal-catalysed cyanation of aryl halides remains one of the most widely adopted and functionally tolerant strategies~\cite{AnbarasanCyanationReview2011}. While substantial methodology development has produced a broad repertoire of candidate catalysts, additives, and cyanide sources, this very diversity introduces three intersecting sources of chemical heterogeneity that challenge traditional modelling approaches. First, both monodentate and bidentate phosphine ligands are competitive across cyanation substrate classes, yet descriptor libraries developed for monophosphines are not directly transferable to bisphosphines (and vice versa) thus fragmenting the feature space. Second, productive cyanation conditions often require additives spanning chemically disparate classes such as elemental reductants (e.g., \ce{Zn}), organic bases (e.g., \ce{NEt3}), and inorganic salts (e.g., \ce{KOAc}), which resist straightforward unification within standard chemical feature spaces. Finally, the choice of cyanide source carries practical implications beyond reactivity. Classical sources such as \ce{NaCN} and \ce{Zn(CN)2} are highly toxic, whereas safer but less reactive alternatives such as \ce{K4[Fe(CN)6]} are increasingly preferred in industrial process settings despite their poor organic solubility, which necessitates biphasic reaction media~\cite{NauthNonToxicCyanide2019,WilsonNiCyanation2025}. Together, these intersecting dimensions of heterogeneity in cyanation reactions pose a distinct challenge for conventional descriptor frameworks, highlighting the value of more flexible, unified representation strategies.

We assembled a design space for the cyanation of methyl 3-bromo-5-fluorobenzoate, comprising 40 commercially available palladium catalysts spanning both mono- and bidentate ligand classes (e.g., \ce{[Pd(tBuXPhos)(allyl)]OTf}, \ce{[Pd(DPPF)Cl2]}), three cyanide sources of varying toxicity (\ce{NaCN}: median lethal dose (mg/kg) $LD_{50}=3.6$, \ce{Zn(CN)2}: $LD_{50}=54$, \ce{K4[Fe(CN)6]}: $LD_{50}=3613$), five additive options (Zn, KOAc, \ce{NEt3}, \ce{K2CO3}, or none), ten solvents, two co-solvent options, and three temperatures (\Cref{fig:pd_cyanation}a). Conditions in which the reaction temperature exceeded the solvent boiling point were removed, giving a final design space of 30,000 conditions. While constructing a bespoke descriptor representation across these heterogeneous components is in principle feasible, doing so would require laborious, trial-and-error feature engineering with no guarantee of success; our approach bypasses this bottleneck by providing a unified, out-of-the-box featurisation directly from textual descriptions, enabling immediate campaign execution without manual descriptor curation. The full search space is provided in Extended Data Figure \ref{fig:cyanation_search_space}. 

\begin{figure*}[h!]
    \centering
    \includegraphics[width=0.8\linewidth]{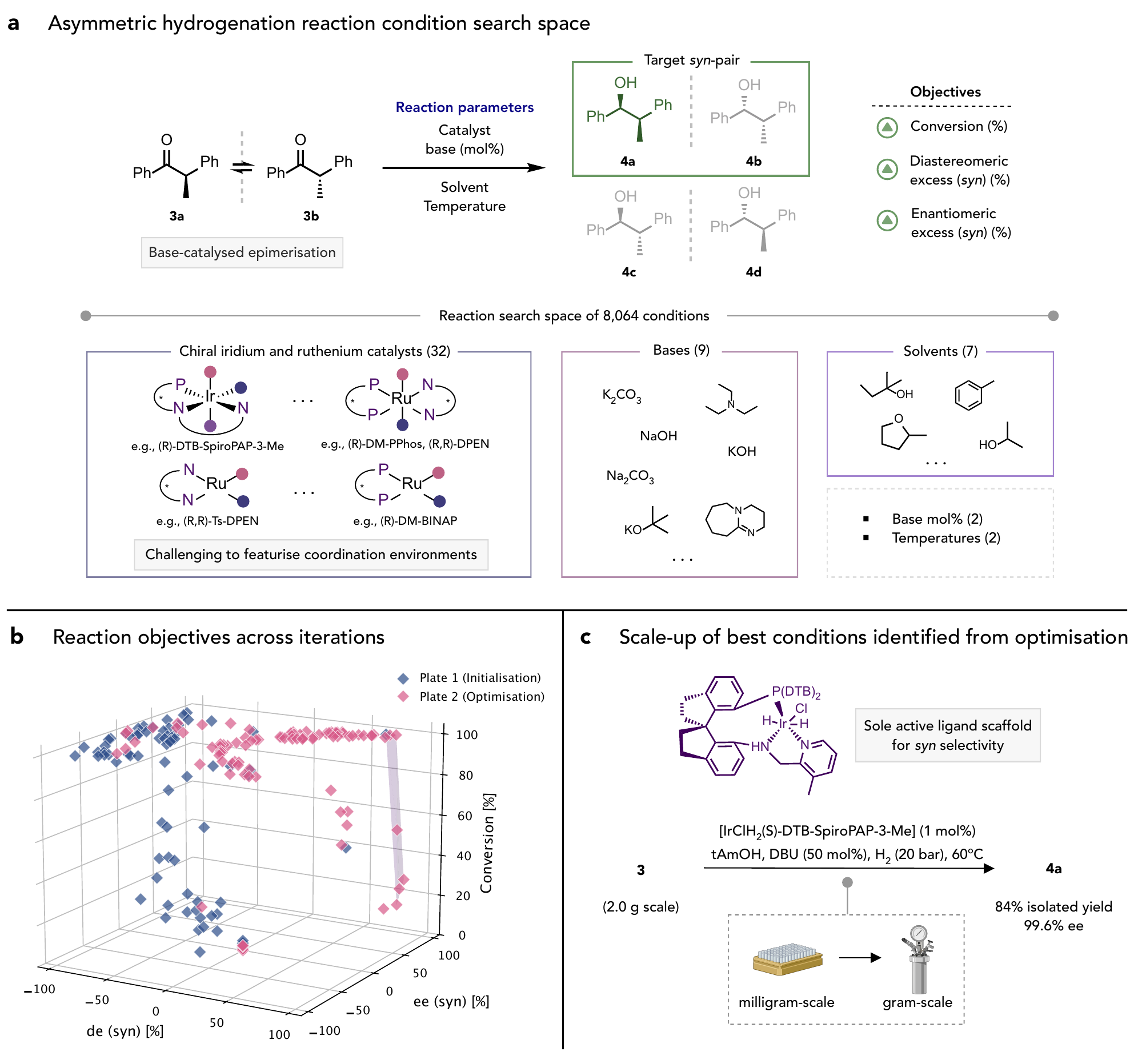}
    \caption{\textbf{Case study 2: Asymmetric ketone hydrogenation.}
\textbf{a,} Search space for the dynamic kinetic hydrogenation of
\textbf{3} to the target \textit{syn}-(1\textit{S},\,2\textit{R}) alcohol \textbf{4a}, one of four accessible stereoisomers
(\textbf{4a}--\textbf{4d}) arising from base-catalysed epimerisation
between \textbf{3a} and \textbf{3b}. Conversion (\%), de~(\textit{syn}) (\%), and ee~(\textit{syn}) (\%) were maximised simultaneously across 32 chiral iridium and ruthenium catalysts, 9 bases, 7 solvents, 2 base
loadings and 2 temperatures, giving 8,064 conditions (full search space in Extended Data Figure \ref{fig:hydrogenation_search_space}).
\textbf{b,} Reaction objectives across HTE optimisation rounds, Plate 1 (Initialisation) and Plate 2   (Optimisation). Positive de and ee denote excess of the target \textit{syn} diastereomer and (1\textit{S},\,2\textit{R}) enantiomer respectively. Negative values denote the corresponding \textit{anti} diastereomer or
(1\textit{R},\,2\textit{S}) enantiomer, respectively. \textit{Syn}-selective outcomes
increased from 5 of 96 conditions in Plate~1 to 59 of 96 in Plate~2.
\textbf{c,} Scale-up of the best conditions identified by optimisation,
delivering \textbf{4a} in 84\% isolated yield and 99.6\% ee at gram scale
(see Supplementary Information Section 3.2). \ce{[IrClH2((R/S)-DTB-SpiroPAP-3-Me)]} was the only catalyst pair in the campaign selective for the target \textit{syn} isomer.}
\label{fig:asymmetric_hydrogenation}
\end{figure*}

Optimisation was initialised with a Sobol-sampled first plate of 96 experiments to probe the global design space. While the first round identified several high-performing conditions, the top hits relied exclusively on toxic cyanide sources (\ce{NaCN} and \ce{Zn(CN)2}), while \ce{K4[Fe(CN)6]} yielded only mediocre outcomes with either poor conversion or selectivity (\Cref{fig:pd_cyanation}b). To test whether high-performing conditions could be identified using a lower-hazard cyanide source, we fixed the cyanide source for round 2 to \ce{K4[Fe(CN)6]} and re-optimised within this constrained domain. The second optimisation round successfully met this target, identifying multiple conditions achieving $>$99\% conversion and $>$99\% selectivity (\Cref{fig:pd_cyanation}b). Notably, high-performing hits spanned both monodentate (\ce{[Pd(tBuXPhos)(allyl)]OTf}) and bidentate (\ce{[Pd(Xantphos)(allyl)]Cl}) ligands, exemplifying the benefit of jointly searching over both ligand families within a unified representation. To confirm that these milligram-scale HTE hits translate to preparative scale-up, we scaled the lead conditions (\ce{[Pd(Xantphos)(allyl)]Cl} with  \ce{K4[Fe(CN)6].3H2O} in DMC/\ce{H2O}) to gram scale (Figure \ref{fig:pd_cyanation}c). The reaction proceeded cleanly, delivering the target aryl nitrile with 94\% isolated yield (see Supplementary Information Section 2). This scalable, low-hazard protocol was identified in just two iterative campaign rounds (one week per round) totalling 192 experiments, evaluating less than 1\% of the 30,000-condition design space.

\subsection{Case study 2: Asymmetric ketone hydrogenation}

Having established the approach on reaction systems with multi-component heterogeneity, we next sought to evaluate its capacity to navigate stereoselective transformations. Catalytic asymmetric hydrogenation of prochiral ketones is among the most widely deployed enantioselective methods in pharmaceutical and agrochemical synthesis~\cite{Blaser2001EnantioSelectiveCatalysis}. In particular, the dynamic kinetic hydrogenation of $\alpha$-substituted aryl ketones, in which two contiguous stereocentres are set in a single step to yield up to four potential stereoisomers~\cite{NoyoriHydrogenation1989,YurinoAsymmetricHydrogenationKetones2024}, exemplifies a class of multi-objective optimisation problems that require balancing trade-offs across a Pareto front defined by conversion, diastereomeric excess (de), and enantiomeric excess (ee). Productive catalysts include chiral iridium and ruthenium complexes drawn from a structurally diverse pool of ligand motifs, spanning bidentate phosphines, mixed phosphine–nitrogen donors (P, N and P, N, N), and a variety of associated ancillary ligands and counter-ions. Representing this coordination diversity in a unified descriptor framework is non-trivial; physical featurisation schemes engineered to quantify the 3D chiral pocket of one ligand geometry or metal centre fail to generalise across fundamentally disparate coordination spheres. Jointly balancing three reaction objectives across this heterogeneous chiral space compounds the modelling challenge. We assembled a design space of 8,064 conditions for the asymmetric dynamic kinetic hydrogenation of 1,2-diphenylpropan-1-one towards the \textit{syn}-(1\textit{S}, 2\textit{R})-1,2-diphenylpropan-1-ol diastereomer, comprising 32 chiral iridium and ruthenium catalysts, 9 organic and inorganic bases, 7 solvents, varied base loading (mol\%), and temperature (\Cref{fig:asymmetric_hydrogenation}a). The full search space is provided in Extended Data Figure \ref{fig:hydrogenation_search_space}.

Optimisation was initialised with a Sobol-sampled first HTE plate of 96 experiments spanning the full design space (Figure \ref{fig:asymmetric_hydrogenation}b). The first round of experiments revealed that the undesired \textit{anti} diastereomer was favoured across the majority of conditions evaluated, yielding multiple hits with high anti selectivity ($-$99\% de (\textit{syn})). In contrast, accessing the target \textit{syn} diastereomer was more challenging: only 5 out of 96 experiments were \textit{syn} selective, and out of the 32 chiral catalysts tested, only a single catalyst pair, \ce{[IrClH2((R/S)-DTB-SpiroPAP-3-Me)]}, demonstrated activity towards the \textit{syn} isomer. The top hit for the desired isomer in Plate 1 achieved quantitative conversion but moderate stereoselectivity (73.5\% de (\textit{syn}) and 86.7\% ee (\textit{syn})) (\Cref{fig:asymmetric_hydrogenation}b). In achiral media, enantiomeric catalyst pairs yield mirror-image products, so each training observation was augmented with its reflected counterpart under a sign inversion of the enantiomeric excess (see Supplementary Information Section 3.3).

Guided by the updated LLM representations, the framework refocused optimisation towards \textit{syn}-producing regions, increasing \textit{syn}-selective outcomes from 5/96 to 59/96 conditions in Plate 2 (Figure \ref{fig:asymmetric_hydrogenation}b). The Pareto front was expanded towards higher stereoselectivity, with the most stereoselective conditions identified reaching 87.1\% de (\textit{syn}) and 93.2\% ee (\textit{syn}). The most balanced lead hit achieved $>$99\% conversion, 80.5\% de (\textit{syn}), and 89.7\% ee (\textit{syn}). Identifying the productive catalyst family did not by itself resolve the optimisation problem: across the reaction conditions employing \ce{[IrClH2((R/S)-DTB-SpiroPAP-3-Me)]}, diastereoselectivity spanned -67\% to +87.1\% de (\textit{syn}), and in 12 instances its sign inverted upon changing temperature or base loading alone, consistent with the dynamic kinetic nature of the transformation. Systematically removing the highest-performing conditions from the Plate 1 training data did not prevent the model from recovering high-performing \textit{syn}-selective conditions, indicating that the outcome did not depend on a small number of fortunate initial hits (see Supplementary Information Section 3.4). Translating the lead conditions identified from optimisation to preparative gram scale proceeded smoothly (Figure \ref{fig:asymmetric_hydrogenation}c): the lead conditions, \ce{[IrClH2((S)-DTB-SpiroPAP-3-Me)]} with \ce{DBU} (50 mol\%) in \ce{tAmOH} at \SI{60}{\celsius}, delivered the target \textit{syn} alcohol with the desired enantiomer in 84\% isolated yield with 99.6\% enantiomeric excess (see Supplementary Information Section 3). This scalable protocol was identified in two iterative campaign rounds (one week per round) using 192 experiments out of 8,064, corresponding to approximately 2.4\% of the design space.

\section{Outlook}
\label{sec:outlook}

This work demonstrates that dynamic language model representations provide a general route to multi-objective reaction optimisation, applicable across chemically diverse reaction systems. By encoding reaction components directly from textual descriptions, the framework searches jointly over chemical entities of different classes (e.g., ligand families, additives, and catalysts) within a single representation, without descriptors being selected or computed for each new system. Applied prospectively to two wet-lab campaigns, it identified conditions that translated directly to preparative scale within two rounds of high-throughput experimentation. Engineered descriptors are chemically interpretable, but their selection presupposes an understanding of which molecular properties govern reactivity in the system at hand. Learning the representation dynamically from text removes this requirement, allowing optimisation to begin before such understanding is established, where mechanistic insight could emerge from the conditions identified rather than being needed to find them. By removing expert-guided featurisation as a prerequisite, we anticipate this approach will extend model-guided optimisation to a wider range of chemistry, including biocatalysis and enzymatic reactions, and ultimately enable more efficient chemical processes.

\section*{Methods}

\subsection*{Metrics for multi-objective optimisation problems}

We consider the simultaneous maximisation of $M$ objective functions  $y_m \colon \mathcal{X} \to \mathbb{R}$, $m = 1, \dots, M$ (e.g., reaction yield and selectivity), over a finite chemical design space $\mathcal{X} = \{\mathbf{x}_1, \dots, \mathbf{x}_N\}$, where each $\mathbf{x}_i$ is a vector of reaction conditions. The aim is to identify the Pareto optimal set, also referred to as the Pareto front. We say that $\mathbf{x}'$ \textit{dominates} $\mathbf{x}$ (written  $\mathbf{x}' \succ \mathbf{x}$) if $y_m(\mathbf{x}') \geq y_m(\mathbf{x})$ for all $m = 1, \dots, M$ and $y_j(\mathbf{x}') > y_j(\mathbf{x})$ for some $j \in \{1, \dots, M\}$. The Pareto optimal set is then defined as:

\begin{equation}\label{eq:pareto}
    \mathcal{P}^{*} = \bigl\{\mathbf{x} \in \mathcal{X} \;\big|\; 
    \nexists\, \mathbf{x}' \in \mathcal{X} : \mathbf{x}' \succ \mathbf{x}\bigr\}
\end{equation}

We seek to approximate $\mathcal{P}^{*}$ using as few experimental iterations or evaluations as possible with multi-objective Bayesian optimisation. The quality of a candidate Pareto front $\mathcal{P} \subseteq \mathbb{R}^M$, obtained from the current set of observations, is measured by its dominated hypervolume (\Cref{fig:figure_2}a) relative to a reference point $\mathbf{r} \in \mathbb{R}^M$, chosen such that $r_m < p_m$ for all $\mathbf{p} \in \mathcal{P}$ and $m = 1, \dots, M$:

\begin{equation}\label{eq:hypervolume}
\begin{split}
    \mathrm{HV}(\mathcal{P}, \mathbf{r}) = \mathrm{Vol}\Bigl( & \bigl\{\mathbf{y} \in \mathbb{R}^M \mid \exists\, \mathbf{p} \in \mathcal{P}: \\
    & r_m \leq y_m \leq p_m \;\forall\, m\bigr\}\Bigr)
\end{split}
\end{equation}

where $\mathrm{Vol}(\cdot)$ denotes the Lebesgue measure. For all benchmark datasets, the optimisation objectives are percentages bounded on $[0, 100]$, and we set $\mathbf{r} = (0, 0)$. The same reference point was used for all featurisation methods. Hypervolume is a widely used quality indicator for multi-objective optimisation, 
rewarding both convergence to the true Pareto front and 
diversity along it~\cite{torresMultiObjectiveActiveLearning2022, ZhangSchottenqNEHVI2024, daultonDifferentiableExpectedHypervolume2020, daultonParallelBayesianOptimization2021}. Importantly, it is monotonic and Pareto-compliant~\cite{guerreiroHypervolumeIndicatorProblems2022a, AudetMultiObjIndicators2021}, meaning that if one candidate Pareto front strictly dominates another, it will achieve a 
strictly higher hypervolume. In this work, we 
report the normalised hypervolume percentage, defined as 
$\mathrm{HV}(\mathcal{P}, \mathbf{r}) / \mathrm{HV}(\mathcal{P}^{*}, \mathbf{r}) 
\times 100\%$, to evaluate the quality of candidate Pareto fronts $\mathcal{P}$ identified by optimisation algorithms relative to 
the true Pareto front $\mathcal{P}^{*}$.

\subsection*{Representing reaction conditions}

We benchmarked three featurisation approaches for representing reaction conditions.

\textbf{One-hot encoding}. As a universally applicable and inexpensive baseline that has shown competitive performance in prior work~\cite{RankoviOHE2024, shieldsBayesianReactionOptimization2021, torresMultiObjectiveActiveLearning2022}, each unique categorical variable (e.g., XPhos, dioxane, PhMe) was represented as a binary indicator vector, and all component vectors were concatenated to form the input representation (\Cref{fig:figure_1}a).

\textbf{Molecular descriptors}. Widely adopted as the primary featurisation strategy in reaction optimisation, molecular descriptors from established descriptor libraries were included as a chemically informative baseline~\cite{daltonUtopiaPointBayesian2024, sinHighlyParallelOptimisation2025, AlphaPSO2026, RomerNickelAlkeneEDBO2024, CadgeNickelDielsEDBO2025, CadgeBisphosphine2025, OcampoMolli2024, ChristensenDataAutonomous2021}. Monophosphine ligands were represented using 190 DFT descriptors from the Kraken library~\cite{genschComprehensiveDiscoveryPlatform2022}, with Principal Component Analysis (PCA)~\cite{MakiewiczPCA1993} applied to retain components explaining 99\% of the variance, reducing dimensionality while preserving information. Solvents were parameterised using four COSMOtherm-derived DFT descriptors from the COSMO-RS database~\cite{moityPanoramaSustainableSolvents2012}. The remaining categorical variables (e.g., bases and precursors), for which comparable descriptor libraries are not readily available, were represented using one-hot encoding.

\textbf{Large language model representations}. Natural language provides a flexible modality for encoding chemically heterogeneous reaction components without requiring component-class-specific featurisation. Each reaction condition $\mathbf{x}_i \in \mathcal{X}$ was represented as a textual prompt of component-value pairs, for example:

\texttt{Reaction condition: ligand: \{ligand name\} solvent: \{solvent name\} precursor: \{precursor name\} base: \{base name\}}

\noindent with one pair per parameter varied in the design space.

A pre-trained language model $h_\phi$ maps the tokenised prompt to a sequence of embedding vectors $\mathbf{H}_i = h_\phi(\mathbf{x}_i) \in \mathbb{R}^{L \times d_{\mathrm{emb}}}$, where $L$ is the sequence length and $d_{\mathrm{emb}}$ is the model's hidden dimension. A pooling operator $\textsc{Pool} \colon \mathbb{R}^{L \times d_{\mathrm{emb}}} \to \mathbb{R}^{d_{\mathrm{emb}}}$ aggregates the token-level representations into a fixed-dimensional embedding $\mathbf{e}_i = \textsc{Pool}(\mathbf{H}_i) \in \mathbb{R}^{d_{\mathrm{emb}}}$ to produce a sequence-level representation. We adopted pooling strategies based on each model's architecture. For encoder-decoder models, only the encoder stack was used, and pooling was applied over its hidden states. The T5 encoder-decoder models (T5-base~\cite{2020t5-basecitation}, T5-small, and T5-chem~\cite{christofidellis2023T5-Chemunifying}) used mean pooling, which averages hidden states over non-padded tokens. We used last-token pooling for the decoder-only model Qwen2.5~\cite{qwen2}, which takes the hidden state at the last non-padding position, as causal attention ensures that only this token has attended to the full input sequence~\cite{radford2018GPT1improving}. Mean pooling was also evaluated for Qwen2.5. For the BART-base encoder-decoder model~\cite{Bartbase}, we used mean pooling following the T5 models, and additionally evaluated CLS pooling as BART inherits a \texttt{<s>} classification token. All models were accessed via the Hugging Face Transformers library~\cite{HugginFaceTransformers}. Ablation studies over pre-trained language model architectures and pooling strategies across all benchmark datasets are presented in Extended Data \Cref{sec:models-and-pooling}.

\subsection*{Overview of optimisation loop}

We initialised our Bayesian optimisation workflows using low-discrepancy Sobol sequences~\cite{burhenneSamplingBasedSobol2011} to provide broad initial coverage of the design space, which were evaluated to form the initial training data. At each subsequent iteration of the optimisation loop, the surrogate model was fitted to all currently observed data, a batch of candidates was selected by optimising the acquisition function over the remaining design space, and the selected experiments were evaluated and added to the training set.

We used the \texttt{qLogNParEGO} acquisition function~\cite{daultonParallelBayesianOptimization2021, LogEI, PilonRoboChem2026}, a log acquisition function that conditions on noisy baseline observations and applies Chebyshev scalarisation to reduce the multi-objective problem into single-objective subproblems. For batch acquisition, candidates were selected in a greedy sequential fashion, with each selection conditioned on previously chosen pending points.

For surrogate models using fixed input representations (one-hot encoding and molecular descriptors), we used an optimised Gaussian process (GP) configuration adapted from EDBO+~\cite{torresMultiObjectiveActiveLearning2022} and Minerva~\cite{sinHighlyParallelOptimisation2025} following standard marginal likelihood optimisation. For our approach using an LLM-based deep kernel surrogate with learned input representations, the GP hyperparameters and language model parameters were optimised jointly as described below. All computational experiments were repeated over 20 random seeds to report statistical variation.

\subsection*{LLM-based deep kernel Gaussian process surrogate}

The LLM first maps tokenised reaction condition prompts to pooled embeddings $\mathbf{e}_i \in \mathbb{R}^{d\mathrm{emb}}$. For each objective $m = 1, \dots, M$, a trainable projection head $g_{\phi_m}$ transforms the language model embedding $\mathbf{e}_i$ to an objective-specific representation $\mathbf{z}_i^{(m)} = g_{\phi_m}(\mathbf{e}_i)$. A Gaussian process (GP) then models the mapping from each projected representation to its corresponding objective value:

\begin{equation}
\begin{split}
    y_m(\mathbf{x}_i) &= f_m(\mathbf{z}_i^{(m)}) + \epsilon_m, \\
    f_m &\sim \mathcal{GP}(\mu_m, k_{\theta_m}), \\
    \epsilon_m &\sim \mathcal{N}(0, \sigma_m^2)
\end{split}
\end{equation}

where $f_m$ is the latent function for objective $m$, modelled as a Gaussian process with constant mean function $\mu_m$ and kernel function $k_{\theta_m}$, and $\epsilon_m$ is Gaussian observation noise with variance $\sigma_m^2$. We use the Mat\'{e}rn-5/2 kernel:

\begin{equation}\label{eq:matern}
\begin{split}
    k_{\theta_m}(\mathbf{z}, \mathbf{z}') &= \sigma_{f,m}^2 \left( 1 + \frac{\sqrt{5}\rho}{\ell_m} + \frac{5\rho^2}{3\ell_m^2} \right) \exp \left( -\frac{\sqrt{5}\rho}{\ell_m} \right), \\
    &\text{where } \rho = \|\mathbf{z} - \mathbf{z}'\|_2
\end{split}
\end{equation}

where $\sigma_{f,m}^2$ is the signal variance, and $\ell_m$ is the lengthscale. Together with the constant mean function $\mu_m$ and noise variance $\sigma_m^2$, these constitute the GP hyperparameters $\theta_m = \{\mu_m, \sigma_{f,m}^2, \ell_m, \sigma_m^2\}$. The kernel matrix for objective $m$ is given by $\mathbf{K}_m = k_{\theta_m}(\mathbf{Z}^{(m)}, \mathbf{Z}^{(m)}) + \sigma_m^2 \mathbf{I}$, where $\mathbf{Z}^{(m)}$ collects the projected embeddings of all observed reaction conditions. The $M$ GPs are queried jointly by the multi-objective acquisition function. The language model $h_\phi$, projection heads $\{g_{\phi_m}\}_{m=1}^{M}$, 
and GP hyperparameters $\{\theta_m\}_{m=1}^{M}$ are jointly optimised by minimising the sum of negative marginal log-likelihoods:

\begin{equation}\label{eq:joint_loss}
    \mathcal{L} = -\sum_{m=1}^{M} \log p(\mathbf{y}_m \mid 
    \mathbf{Z}^{(m)}, \theta_m),
\end{equation}

where the marginal log-likelihood for objective $m$ is:
\begin{equation}\label{eq:loglikelihood}
\begin{split}
    \log p(\mathbf{y}_m \mid \mathbf{Z}^{(m)}, \theta_m) = 
    &-\frac{1}{2} \biggl[ \tilde{\mathbf{y}}_m^\top \mathbf{K}_m^{-1} \tilde{\mathbf{y}}_m \\
    &+ \log|\mathbf{K}_m| + n\log2\pi \biggr]
\end{split}
\end{equation}
where $n$ is the number of observations and 
$\tilde{\mathbf{y}}_m = \mathbf{y}_m - \mu_m\mathbf{1}$ denotes the targets 
centred by the constant mean function. GP targets are standardised to zero mean and unit variance for training. Gradients from all $M$ objectives are backpropagated through the kernel and projection heads to the shared LLM parameters $\phi$, where each projection head $g_{\phi_m}$ receives gradients only from its corresponding objective. This enables the LLM to learn representations that are jointly optimised for GP predictive performance across all objectives, while each projection head specialises for its target objective.

\subsection*{Parameter-efficient fine-tuning of language models}

The language model embeddings of reaction conditions evolve across BO iterations as the LLM is fine-tuned on accumulating observations. To do so efficiently, we apply  parameter-efficient fine-tuning (PEFT) with Low-Rank Adaptation (LoRA)~\cite{LoRA2021}. Rather than updating all of the pre-trained language model's parameters, we use LoRA to update only a small subset. For a pre-trained weight matrix $\mathbf{W}_0 \in \mathbb{R}^{d \times k}$, LoRA injects trainable low-rank decompositions such that the weight update takes the form:
\begin{equation}\label{eq:lora}
\begin{split}
    \mathbf{W} &= \mathbf{W}_0 + \Delta\mathbf{W}, \\
    \Delta\mathbf{W} &= \frac{\alpha}{r}\mathbf{B}\mathbf{A}, \quad \text{with } \mathbf{B} \in \mathbb{R}^{d \times r}, \mathbf{A} \in \mathbb{R}^{r \times k}
\end{split}
\end{equation}
 where $r \ll \min(d, k)$ is the rank and $\alpha$ is a scaling hyper-parameter. We use LoRA to target a subset of linear projection matrices within the pre-trained language models, using rank $r=4$ and $\alpha=16$.

\subsection*{Computational implementation}

All models, Bayesian optimisation workflows, and computational analyses were implemented in Python 3.10. We used \texttt{PyTorch}~\cite{paszkePyTorchImperativeStyle2019} (v2.8.0), \texttt{GPyTorch}~\cite{gardnerGPyTorchBlackboxMatrixmatrix2018} (v1.14), and \texttt{BoTorch}~\cite{balandatBoTorchFrameworkEfficient2020} (v0.13.0) to construct our multi-objective Bayesian optimisation algorithms. Pre-trained language models were loaded using Hugging Face \texttt{transformers}~\cite{HugginFaceTransformers} (v4.51.2), with Low-Rank Adaptation (LoRA) implemented using \texttt{peft} (v0.15.1) for parameter-efficient fine-tuning. We used \texttt{PyTorch Lightning}~\cite{falconPyTorchLightningPytorchlightning0762020} (v2.5.4) for reproducible random seed management and workflow control. \texttt{Scikit-learn}~\cite{pedregosaScikitlearnMachineLearning2011} (v1.7.1) and \texttt{NumPy}~\cite{harris2020Numpyarray} (v1.26.4) were utilised for machine learning pipelines and numerical data processing, including Principal Component Analysis (PCA) for descriptor dimensionality reduction. Experiment tracking and logging were conducted using \texttt{Weights \& Biases} (\texttt{wandb}, v0.21.0)~\cite{biewaldWANDBexperiment}. \texttt{Matplotlib}~\cite{hunterMatplotlib2DGraphics2007} (v3.10.1) and \texttt{seaborn}~\cite{waskomSeabornStatisticalData2021} (v0.13.2) were used for plotting and visualising all computational results presented in this work. All computations were run on the Roche HPC cluster, using NVIDIA A100 Tensor Core and Blackwell GPUs.

\subsection*{Experimental procedures}

High-throughput experimentation procedures, analytical methods, and scale-up syntheses and characterisation for the palladium-catalysed cyanation and asymmetric ketone hydrogenation reactions are provided in the Supplementary Information.

\section*{Data availability}
All benchmark datasets, reaction condition search spaces, and high-throughput experimentation (HTE) experimental data generated in this study are included in the manuscript, Supplementary Information, and on the accompanying public GitHub repository.

\section*{Code availability}
The custom code developed for this work is implemented in Python and is made available in a public GitHub repository under the Apache 2.0 licence: https://github.com/schwallergroup/alice

\section*{Acknowledgements}
J.W.S., R.P.B., K.P. and R.B. thank Roche and its Technology Innovation and Science (TIS) initiative for financial support. We thank the Global Internship Program in Innovation \& Sustainability (IP2TIS) 2025 (\url{https://careers.roche.com/global/en/ip2tis-program}) for financial support of this project. D.M.S., B.R., and P.S. acknowledge support from NCCR Catalysis (grant no. 225147), a National Centre of Competence in Research funded by the Swiss National Science Foundation. D.M.S. was funded by the Swiss National Science Foundation (SNSF) [226509].

\section*{Author contributions}
J.W.S., B.R., R.B. and P.S. conceived the project. J.W.S., D.M.S. and B.R. developed the computational workflow with input from S.L.C., A.A., R.P.B. and P.S. J.W.S. and D.M.S. performed the computations and trained the models. J.W.S., M.J.N. and R.B. carried out the experimental work. J.W.S., K.P., M.J.N. and R.B. designed and planned the experimental case studies, with input from M.D.R.L. J.W.S., D.M.S. and B.R. analysed the results with help from M.D.R.L., K.P., M.J.N., R.B. and P.S. S.L.C., A.A., R.P.B., K.P., M.J.N., R.B. and P.S. provided supervision. J.W.S. wrote the paper with input from all authors.

\section*{Competing interests}
J.W.S., D.M.S., M.D.R.L., A.A., R.P.B., K.P., M.J.N. and R.B. declare potential financial and non-financial conflicts of interest as full employees of F. Hoffmann-La Roche Ltd. The other authors declare no competing interests.

\newpage

\bibliography{references}
\bibliographystyle{unsrtnat}


\newpage
\appendix
\onecolumn

\setcounter{figure}{0}
\renewcommand{\thefigure}{\arabic{figure}}

\section{Supplemental main text benchmarking studies}
\label{SI_sec:main_text_supp}

\begin{extendeddata*}[h!]
    \centering
    \includegraphics[width=1\linewidth]{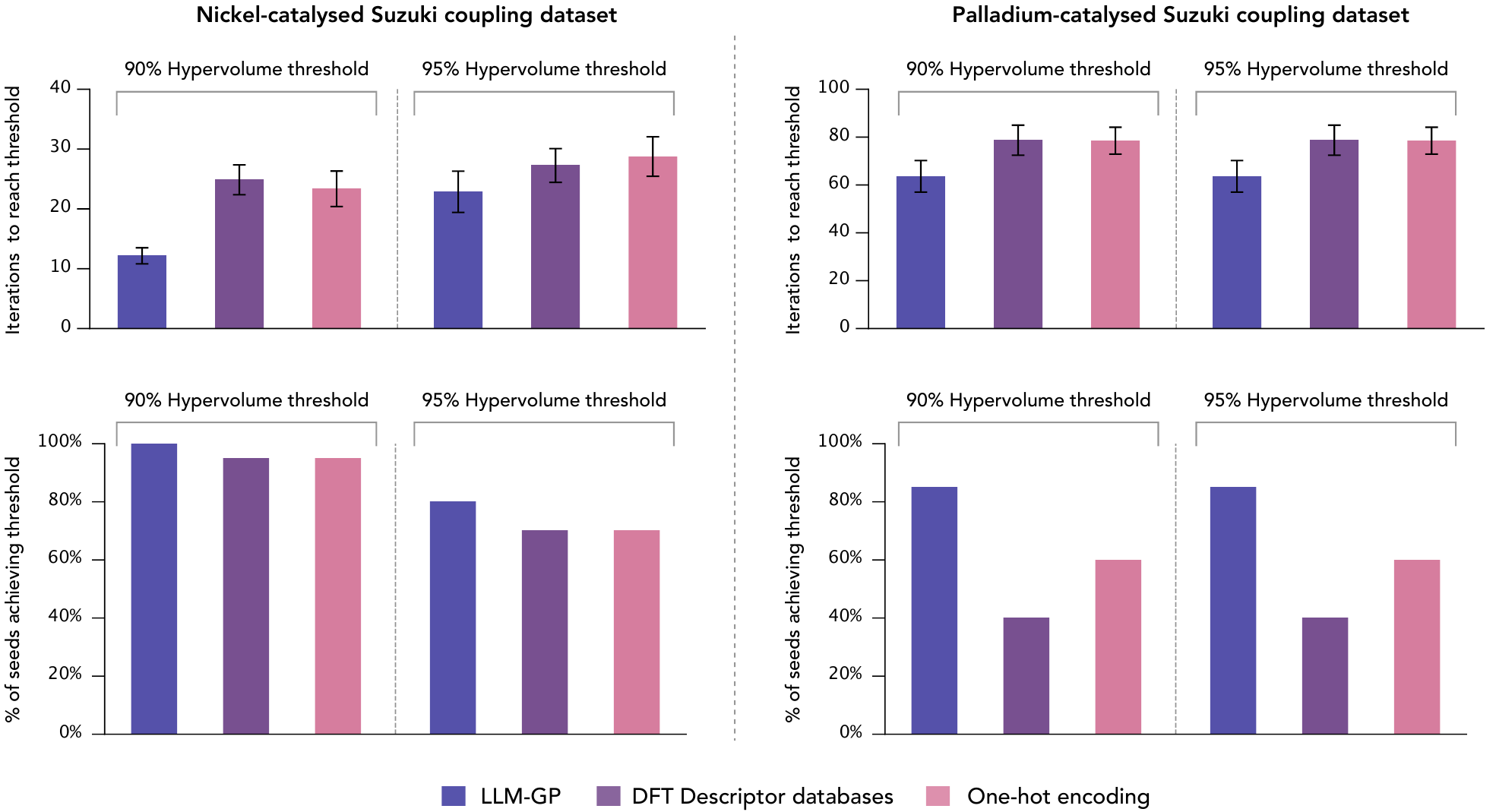}
    \caption{Sequential low-data optimisation benchmarks comparing different representation methods (the LLM-GP framework, DFT descriptor databases, and one-hot encoding) on the nickel-catalysed and palladium-catalysed Suzuki coupling datasets. Top panels: mean number of optimisation iterations required to reach 90\% and 95\% hypervolume thresholds (lower is better, error bars indicate standard error across 20 random seeds). Bottom panels: percentage of optimisation runs (initialised with different random seeds) reaching each threshold within the allocated experimental budget (higher is better).}  
\end{extendeddata*}

\begin{extendeddata*}[h!]
    \centering
    \includegraphics[width=1\linewidth]{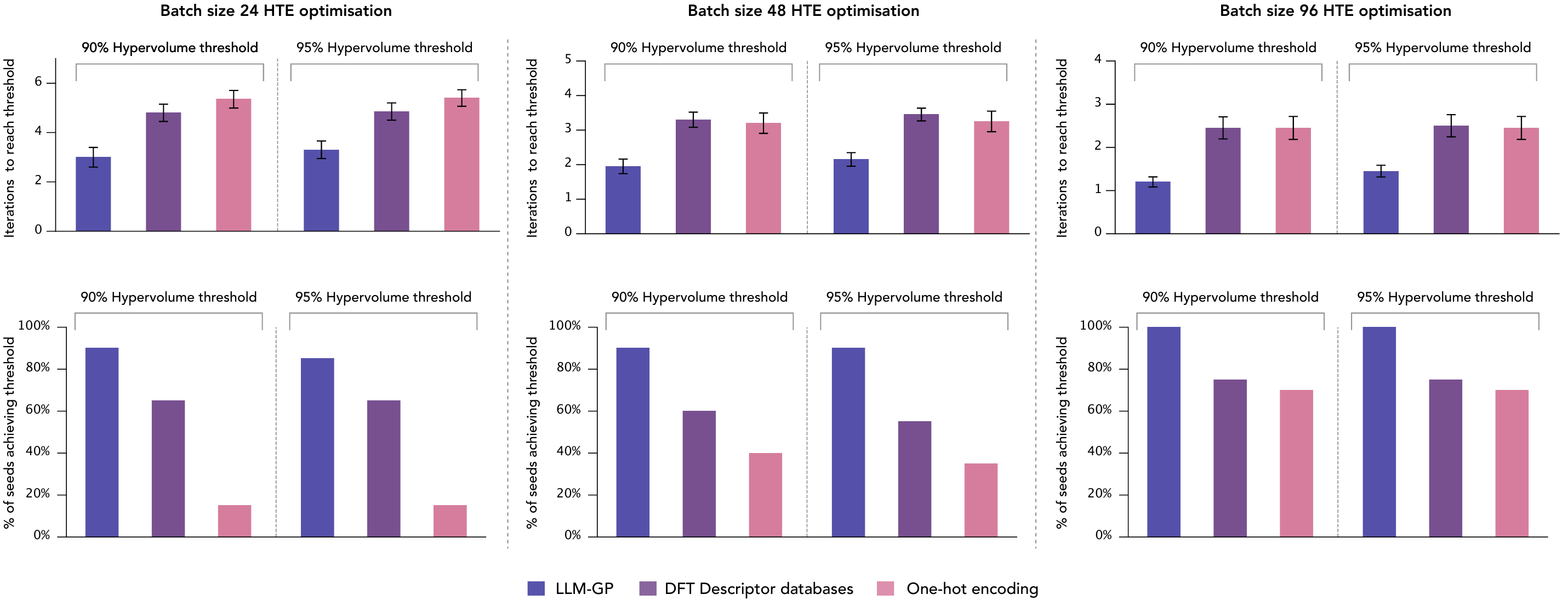}
    \caption{High-throughput experimentation (HTE) optimisation benchmarks on the palladium-catalysed sulfonamide coupling dataset, comparing the LLM-GP framework, DFT descriptor databases, and one-hot encoding across 24-well (left), 48-well (middle), and 96-well (right) plate batch sizes. Top panels: mean number of plate iterations required to reach 90\% and 95\% hypervolume thresholds (lower is better, error bars indicate standard error across 20 random seeds). Bottom panels: percentage of optimisation runs (initialised with different random seeds) reaching each threshold within the allocated experimental budget (higher is better).}
\end{extendeddata*}

\clearpage
\section{Influence of different pre-trained models and pooling strategies}
\label{sec:models-and-pooling}

\begin{extendeddata*}[h!]
    \centering
    \includegraphics[width=0.9\linewidth]{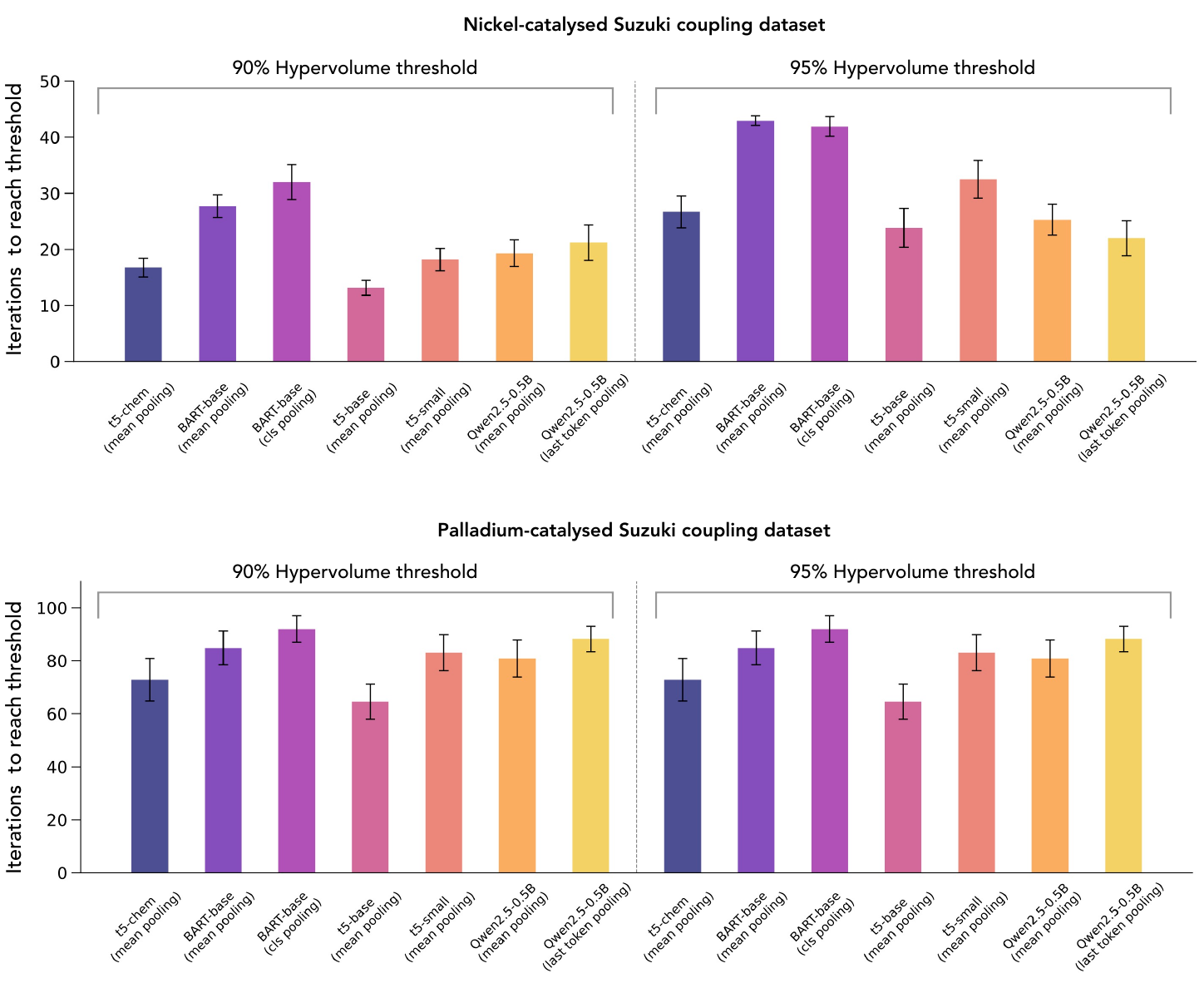}
    \caption{Ablation studies over pre-trained language model architectures and pooling strategies on the nickel-catalysed (top) and palladium-catalysed (bottom) Suzuki coupling datasets. Mean number of optimisation iterations required to reach 90\% and 95\% hypervolume thresholds are compared across T5-chem (mean pooling), BART-base (mean and CLS pooling), T5-base and T5-small (mean pooling), and Qwen2.5-0.5B (mean and last token pooling). Error bars indicate standard error across 20 random seeds. T5-base with mean pooling consistently achieved practical hypervolume thresholds within the fewest iterations across all datasets and thresholds.}
\label{figure-pd-api_suzuki_threshold}
\end{extendeddata*}

\begin{extendeddata*}[h!]
    \centering
    \includegraphics[width=0.9\linewidth]{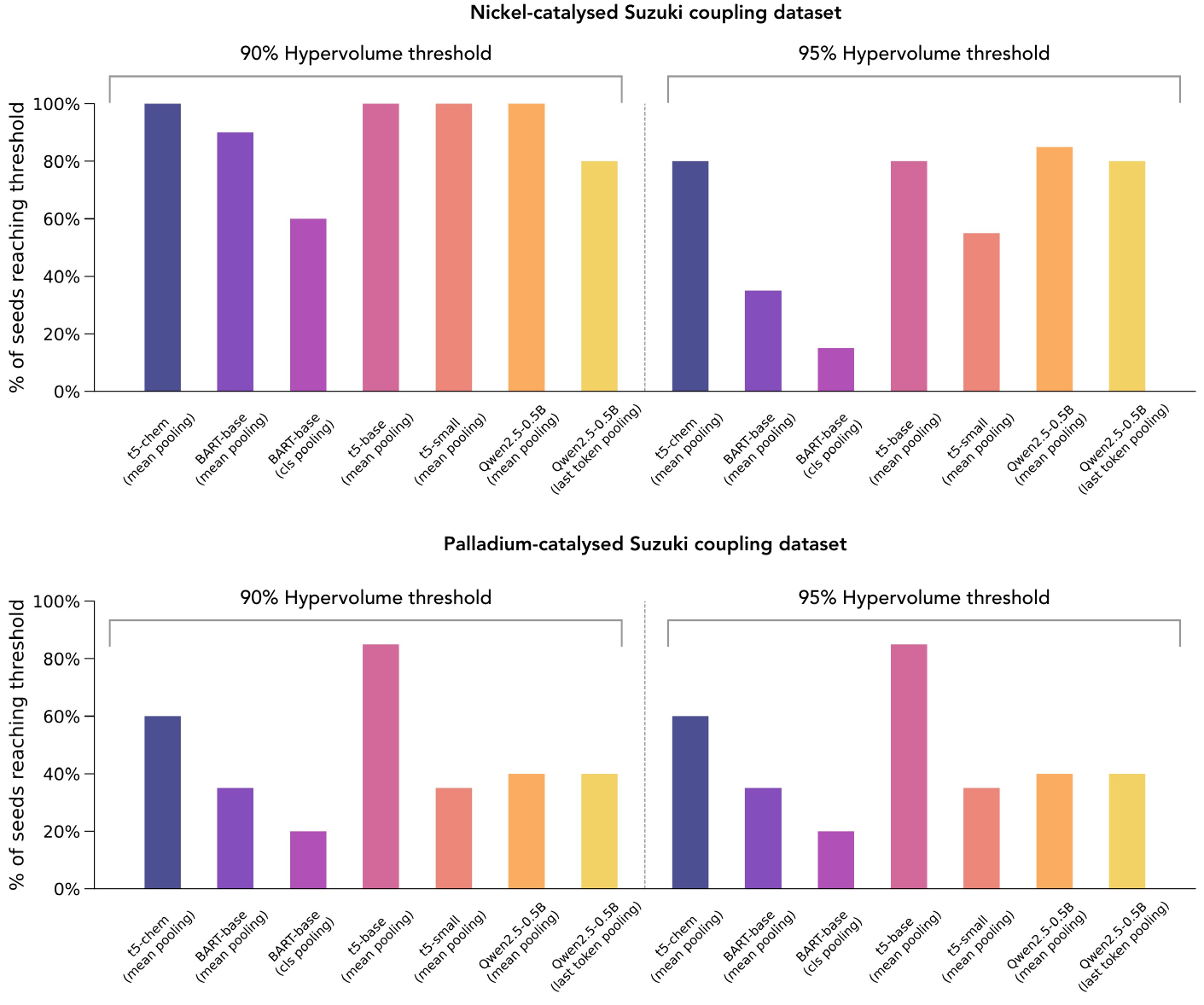}
    \caption{Percentage of optimisation runs (initialised with 20 different random seeds) reaching 90\% and 95\% hypervolume thresholds within the allocated experimental budget, across pre-trained language model architectures and pooling strategies on the nickel-catalysed (top) and palladium-catalysed (bottom) Suzuki coupling datasets. Models and pooling strategies are as described in the Methods.}
\end{extendeddata*}

\begin{extendeddata*}[h!]
    \centering
    \includegraphics[width=0.9\linewidth]{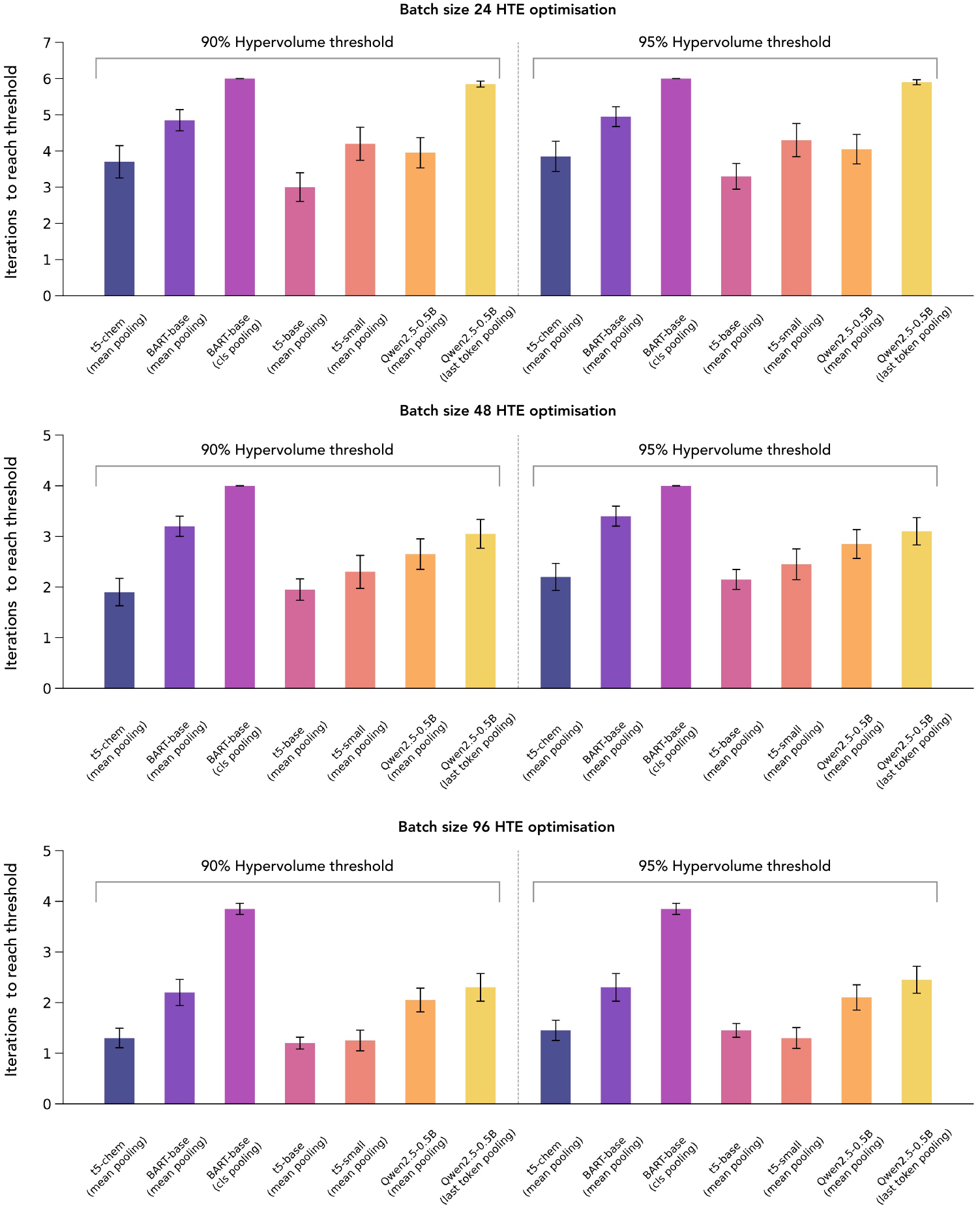}
    \caption{Ablation studies over pre-trained language model architectures and pooling strategies for high-throughput experimentation (HTE) optimisation on the palladium-catalysed sulfonamide coupling dataset, across 24-well (top), 48-well (middle), and 96-well (bottom) plate batch sizes. The bar charts show the mean number of plate iterations required to reach 90\% and 95\% hypervolume thresholds. Models and pooling strategies are as described in the Methods. Error bars indicate standard error across 20 random seeds.}
\end{extendeddata*}

\begin{extendeddata*}[h!]
    \centering
    \includegraphics[width=0.9\linewidth]{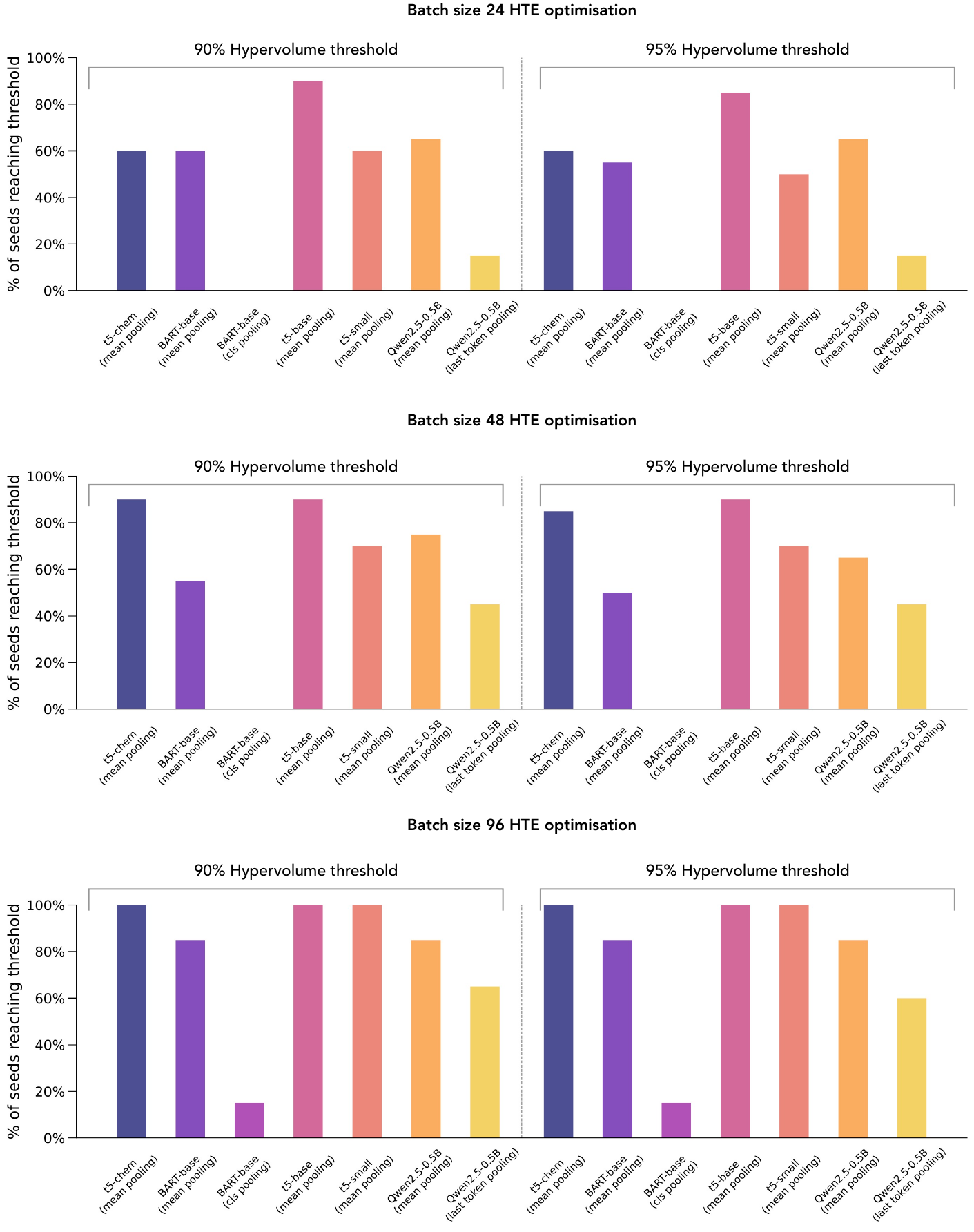}
    \caption{Percentage of optimisation runs (initialised with 20 different random seeds) reaching 90\% and 95\% hypervolume thresholds within the allocated experimental budget, across pre-trained language model architectures and pooling strategies for high-throughput experimentation (HTE) optimisation benchmarks on the palladium-catalysed sulfonamide coupling dataset, across 24-well (top), 48-well (middle), and 96-well (bottom) plate batch sizes. Models and pooling strategies are as described in the Methods.}
\end{extendeddata*}

\clearpage

\section{Design spaces for prospective optimisation campaigns}

\begin{extendeddata*}[h!]
\centering
\includegraphics[width=\textwidth,height=\textheight,keepaspectratio]{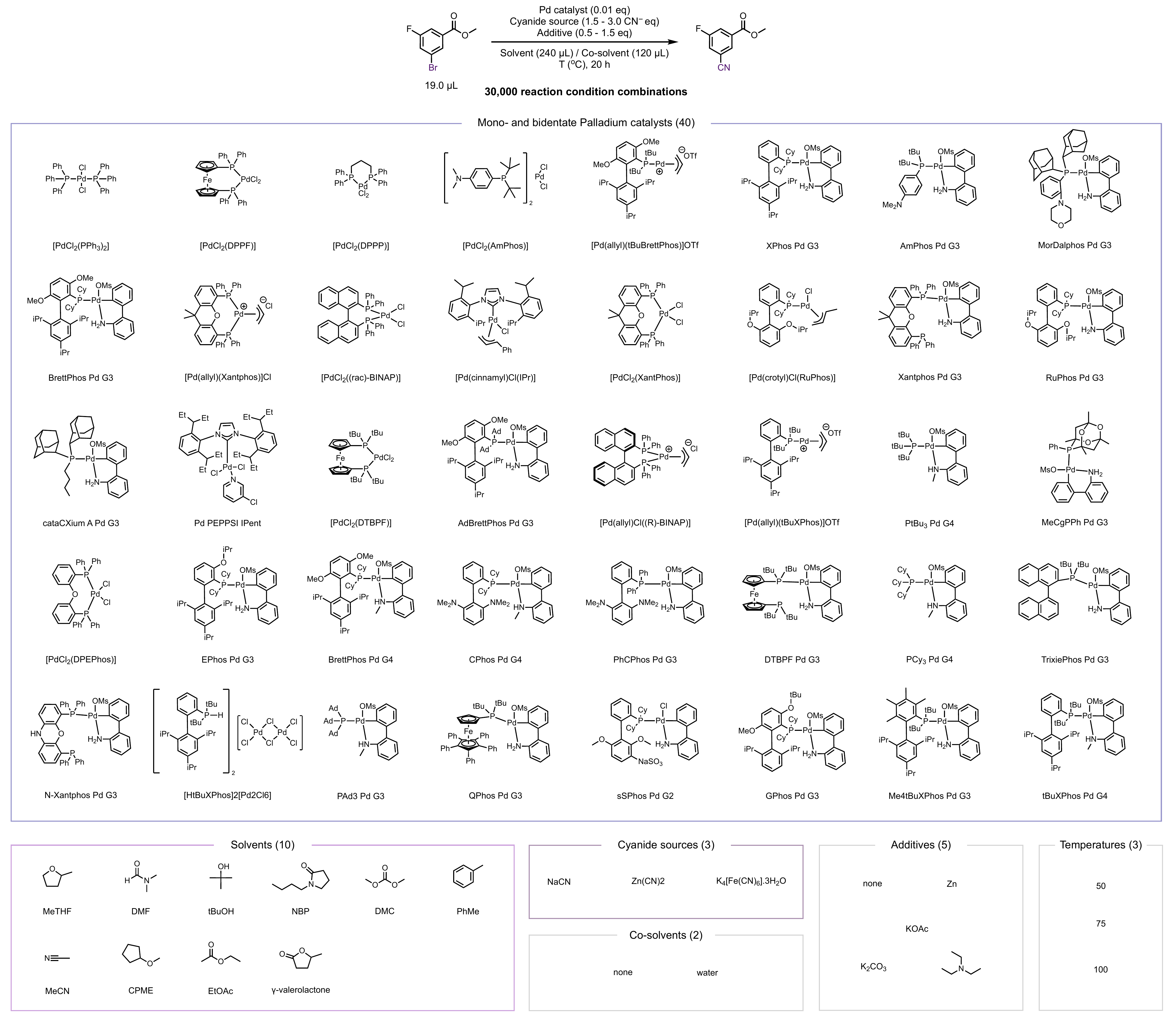}
  \caption{Reaction condition search space for the prospective Pd-catalysed cyanation reaction comprising 40 mono- and bidentate Pd catalysts, 10 solvents, 3 cyanide sources, 2 co-solvents, 5 additives, and 3 temperatures.}
  \label{fig:cyanation_search_space}
\end{extendeddata*}

\begin{extendeddata*}[h!]
\centering
\includegraphics[width=\textwidth,height=\textheight,keepaspectratio]{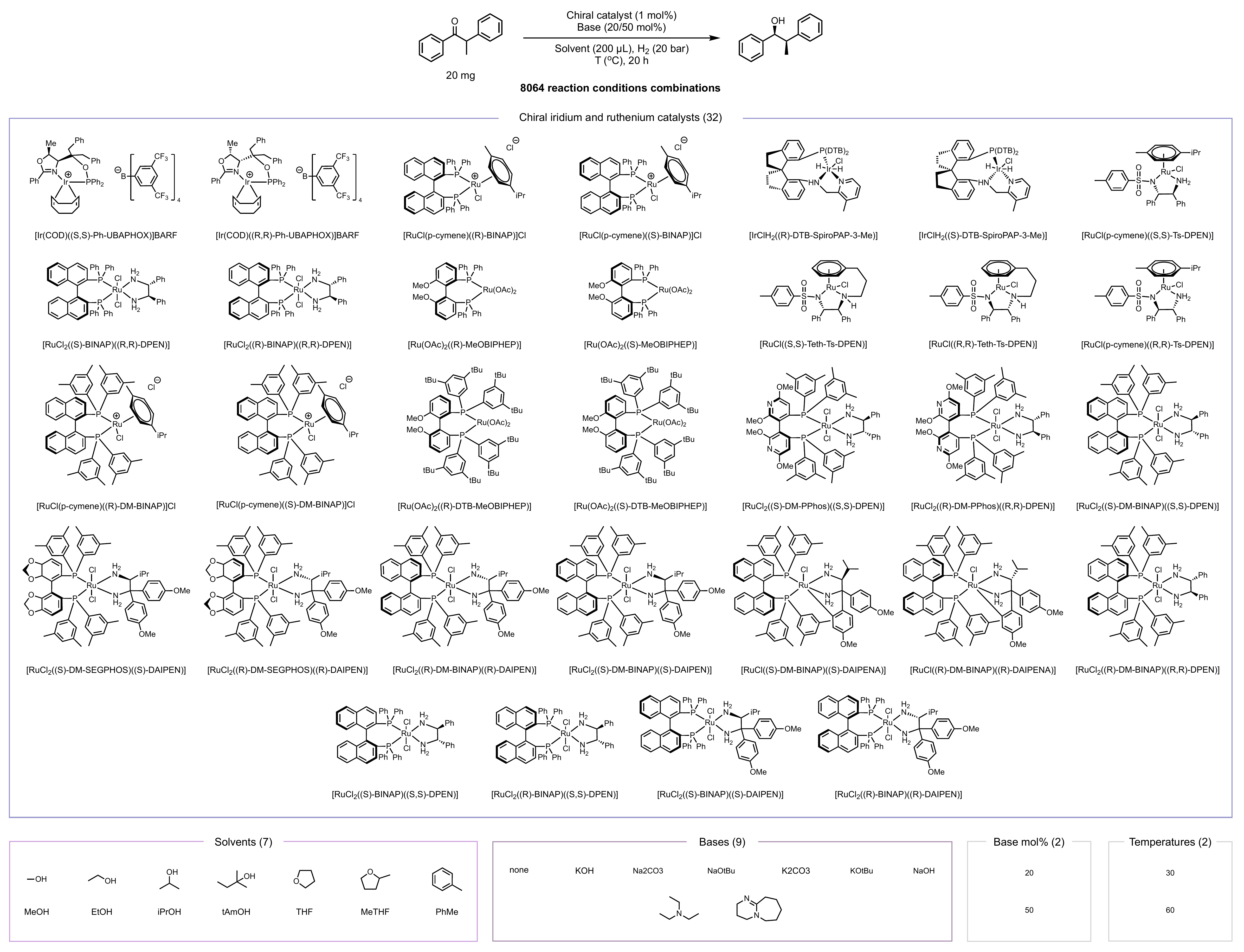}
  \caption{Reaction condition search space for the prospective asymmetric ketone hydrogenation comprising 32 chiral catalysts, 7 solvents, 9 bases, 2 base mol\%, and 2 temperatures.}
  \label{fig:hydrogenation_search_space}
\end{extendeddata*}

\end{document}


\maketitle

\vspace{-3.5em}
\begin{center}
\dag These authors contributed equally to this work.\\
*Corresponding authors. Email:
\email{philippe.schwaller@epfl.ch}
\end{center}

\tableofcontents
\newpage

\section{High-throughput experimentation (HTE) platform}

\begin{figure}[H]
\centering
\includegraphics[width=\textwidth,height=\textheight,keepaspectratio]{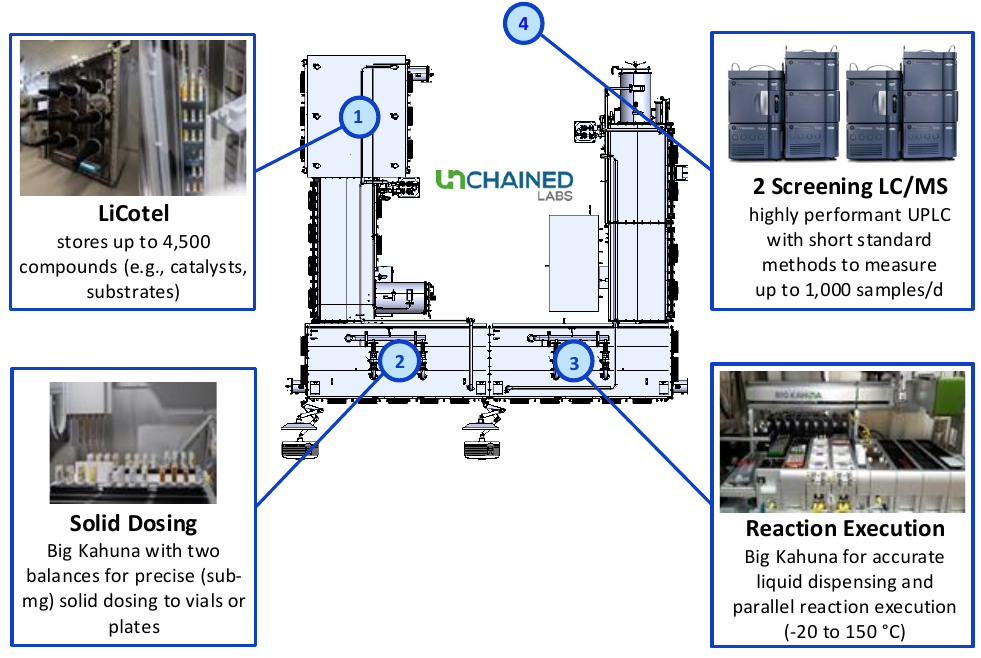}
  \caption{Schematic overview of the high-throughput experimentation (HTE) set-up.}
  \label{fig:hte_setup}
\end{figure}

We conducted all high-throughput experiments using a parallel experimentation platform custom-designed by UnchainedLabs (\Cref{fig:hte_setup}). This system comprises two interconnected Big Kahuna platforms, one of which is further integrated with a LiCONiC LiCotel system. The entire setup is encased within an LC Technology Solutions glove box featuring dual circulation systems, with one dedicated to solid dispensing and another to reaction execution. LC-MS analysis was performed using an ACQUITY UPLC I-Class system with QDa from Waters. For Chiral HPLC analyses, a Chiralcel OJ-3 column from Daicel was used, with ethanol and heptane as the mobile phase.\\

Solid components were dispensed with a Mettler balance in vial dispense mode, using SV hoppers for precursors and ligands and SV hoppers ($\le$ 15 mg) or 10 mL classic hoppers ($>$15 mg) for solid additives. For target dispense quantities $<$ \SI{0.4}{\milli\gram}, materials were dispensed as coated ChemBeads. Following automated solid dispensing, liquid components (substrates, liquid additives, and solvents) were manually added using an Eppendorf Multipette E3 single channel pipette (4987000010). Reactions were performed in standard 96-position parallel synthesis reaction blocks (Analytical Sales and Services, SKU: 96960), with V\&P Scientific super tumble stir discs (VP 721F-1) for stirring, or in a Screening Pressure Reactor (SPR) from UnchainedLabs. Data analysis was performed using the HTE OS workflow described by Wuitschik et al.~\cite{Wuitschik2024} HTE OS is integrated with a SpotFire application that enables tagging of LC-MS and HPLC signals into categories (e.g., “limiting SM”, “other SM”, “solvent”, “ignore peak”). During analysis, all peaks were tagged accordingly, with peaks corresponding to ligands and precatalyst components designated as “ignore peak”. For LCAP calculations, signals tagged as “other SM”, “solvent”, or “ignore peak” were excluded from consideration.

\newpage

\section{Prospective application: Palladium-catalysed cyanation}

\subsection{HTE experimental procedure}

For all HTE plates, the following experimental procedure was performed. Palladium catalysts (1 mol\%), cyanide sources, and solid additives were dispensed into \SI{1}{\milli\liter} vials with stirring disks in a 96-well plate. Methyl 3-bromo-5-fluorobenzoate (19.0 \si{\micro L}, 129 \si{\micro mol}), solvents (240 \si{\micro L}), and cosolvent (120 \si{\micro L}) were added to the vials, followed by the addition of liquid additive \ce{NEt3} (26.9 \si{\micro L}, 1.50 eq) where applicable. The plate was sealed and stirred (400 rpm) at the specified temperature for 20 h. 4/1 \ce{MeCN}/\ce{H2O} (400 \si{\micro L}) was added to each well and shaken for \SI{20}{\minute} at room temperature. Samples were taken and analysed by Liquid Chromatography-Mass Spectrometry (LC-MS).

\subsection{Scale-up and synthesis of methyl 3-cyano-5-fluorobenzoate}

\begin{figure}[H]
\centering
\includegraphics[width=0.9\textwidth,height=\textheight,keepaspectratio]{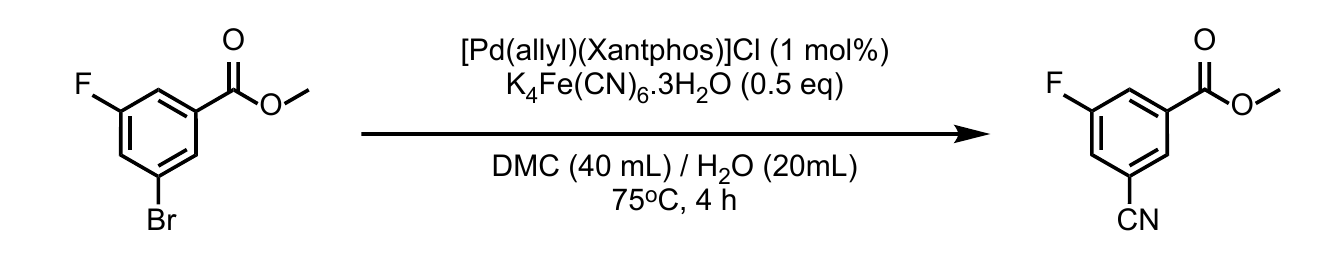}
  \label{fgr:cyanation_scale_up}
\end{figure}

A nitrogen-flushed 100 mL Easymax flask with overhead stirring was charged with methyl 3-bromo-5-fluorobenzoate (3.17 mL, 21.5 mmol) and DMC (40 mL). In a glovebox, a solution of \ce{K4Fe(CN)6.3H2O} (4.53 g, 10.7 mmol, 0.50 eq) in water (20 mL) was prepared and added to the Easymax flask under argon. The mixture was heated to \SI{75}{\degreeCelsius}. Once the temperature was reached, \ce{[Pd(allyl)(Xantphos)]Cl} (163.4 mg, 215 \si{\micro mol}, 1.0 mol\%) was added and the mixture was stirred at 75 °C for 4 h. The reaction was transferred with degassed 25\% aq. \ce{NaCl} solution (100 mL) to a separation funnel and extracted three times with degassed \ce{EtOAc} (3 $\times$ 100 mL). The combined organic phases were washed with sat. aq. NaCl solution, dried over \ce{Na2SO4}, filtered, and the solvent was removed using a rotary evaporator. The crude product was purified by flash column chromatography on silica gel (\ce{EtOAc}:\ce{heptane} = 0:100 to 30:70) over 20 min, 80 g, only UV active fractions collected), affording the product as a white solid. Yield: 3.61 g (94\%). \\

\textbf{\textsuperscript{1}H NMR} (400 MHz, \ce{CDCl3}): $\delta$ 8.18 – 8.13 (\textit{m}, 1H, Ar\textit{H}), 8.00 – 7.95 (\textit{m}, 1H, Ar\textit{H}), 7.58 – 7.53 (\textit{m}, 1H, Ar\textit{H}), 3.98 (\textit{s}, 3H, OC\textit{H}\textsubscript{3}).\\

\textbf{\textsuperscript{13}C\{\textsuperscript{1}H\} NMR} (101 MHz, \ce{CDCl3}): $\delta$ 164.0 (\textit{d}, \textsuperscript{4}$J_\text{F,C}$ = 2.9 Hz, \textit{C}O\textsubscript{2}Me), 162.1 (\textit{d}, \textsuperscript{1}$J_\text{F,C}$ = 252.4 Hz, arom.), 133.8 (\textit{d}, \textsuperscript{3}$J_\text{F,C}$ = 7.7 Hz, arom.), 129.2 (\textit{d}, \textsuperscript{4}$J_\text{F,C}$ = 3.7 Hz, arom.), 123.1 (\textit{d}, \textsuperscript{2}$J_\text{F,C}$ = 24.6 Hz, arom.), 121.4 (\textit{d}, \textsuperscript{2}$J_\text{F,C}$ = 23.1 Hz, arom.), 116.7 (\textit{d}, \textsuperscript{4}$J_\text{F,C}$ = 2.9 Hz, \textit{C}N), 114.3 (\textit{d}, \textsuperscript{3}$J_\text{F,C}$ = 8.8 Hz, arom.), 53.0 (O\textit{C}H\textsubscript{3}).

\begin{figure}[H]
\centering
\includegraphics[width=\textwidth,height=\textheight,keepaspectratio]{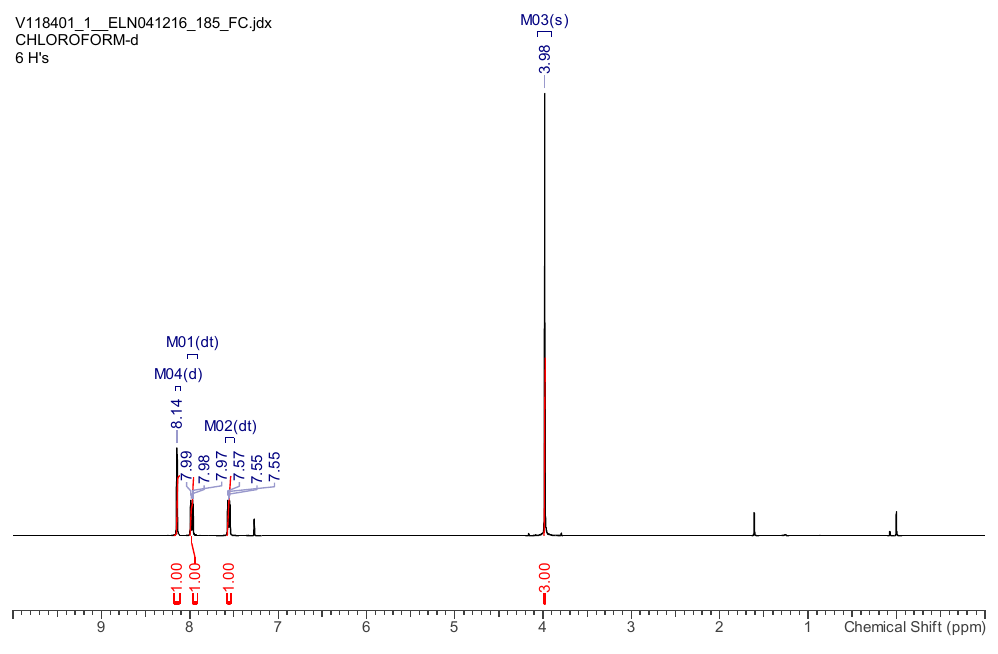}
  \caption{${^1}$H NMR (400 MHz, \ce{CDCl3}) of methyl 3-cyano-5-fluorobenzoate}
  \label{fgr:HNMR_cyanation}
\end{figure}

\begin{figure}[H]
\centering
\includegraphics[width=\textwidth,height=\textheight,keepaspectratio]{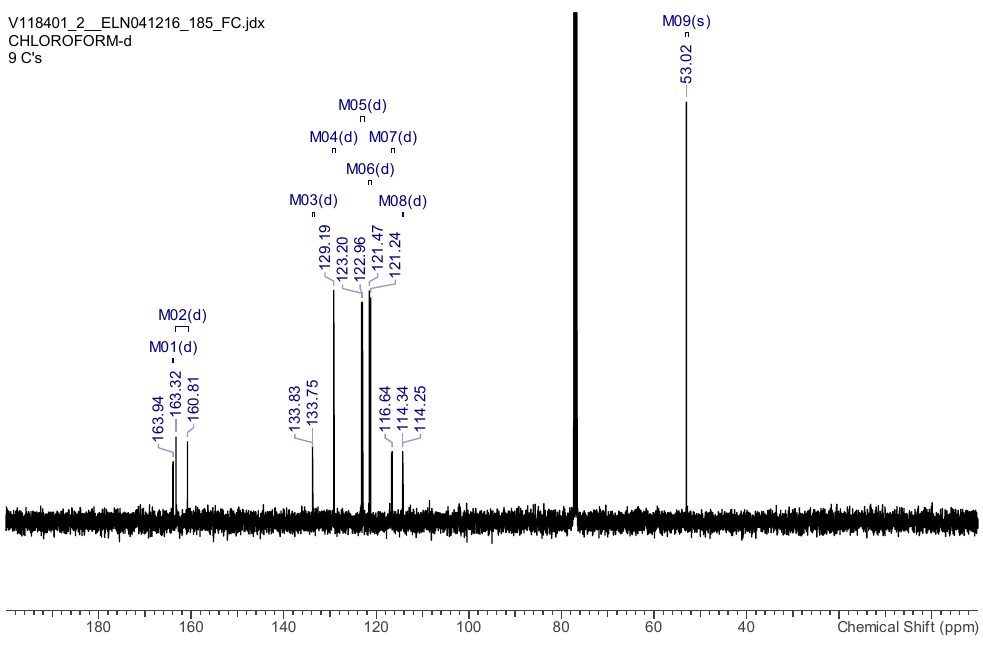}
  \caption{${^{13}}$C\{${^1}$H\} NMR (101 MHz, \ce{CDCl3}) of methyl 3-cyano-5-fluorobenzoate}
  \label{fgr:CNMR_cyanation}
\end{figure}

\newpage

\section{Prospective application: Asymmetric ketone hydrogenation}

\subsection{HTE experimental procedure}

For all HTE plates, the following experimental procedure was performed inside a nitrogen filled glovebox. Iridium and ruthenium catalysts (1 mol\%) and solid bases were dispensed into \SI{1}{\milli\liter} vials in a 96-well plate. Stock solutions of 1,2-diphenylpropan-1-one (20 mg, 95.1 \si{\micro mol} in solvent (200 \si{\micro L})) were prepared and added to the vials, followed by addition of liquid bases, where applicable. The plate was sealed inside the glovebox, placed in a screening pressure reactor (SPR) and shaken overnight at the indicated temperature under 20 bar \ce{H2}. The solvent was removed using a GeneVac and the residue redissolved in EtOH (200 \si{\micro L}). Samples were taken and analysed by chiral HPLC.

\subsection{Scale-up and synthesis of (1\textit{S}, 2\textit{R})-1,2-diphenylpropan-1-ol}

\begin{figure}[H]
\centering
\includegraphics[width=\textwidth,height=\textheight,keepaspectratio]{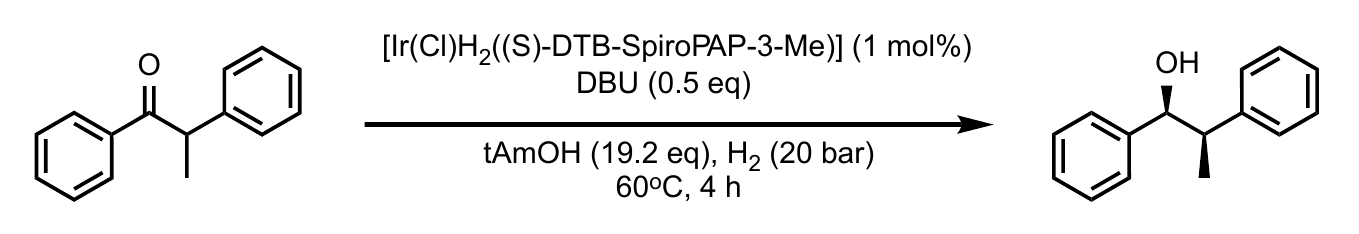}
  \label{fgr:hydrogenation_scale_up}
\end{figure}

In an argon filled glovebox, \ce{[Ir(Cl)H2((S)-DTB-SpiroPAP-3-Me)]}  (93.1 mg, 95.1 \si{\micro mol}, 0.01 eq) and 1,2-diphenylpropan-1-one  (2000 mg, 9.51 mmol, 1.0 eq) were added to a 50 mL autoclave. Degassed \ce{tAmOH}  (16.1 g, 20 mL, 19.2 eq)  and DBU  (724.0 mg, 711.2 \si{\micro L}, 4.76 mmol, 0.5 eq)  were subsequently added to the reaction mixture. The autoclave was then sealed, pressurised with Ar (5 bar) and removed from the glove box. The autoclave was connected to a \ce{H2} line, purged five times with \ce{H2} (10 bar) and stirred at 60 °C and 500 rpm under 20 bar of \ce{H2}. Reaction progress was monitored via \ce{H2} uptake, reaching completion within 4 h. The autoclave was allowed to cool to room temperature and carefully depressurised. The deep yellow reaction mixture was transferred with \ce{EtOAc} (130 mL) to a separation funnel and washed sequentially with 1 M aqueous HCl solution (2 * 50 mL), water (50 mL) and saturated aq. NaCl solution (50 mL). The organic phase was dried over \ce{Na2SO4}, filtered, and concentrated under reduced pressure to afford a lightly brown oil. 
\\

This crude oil was purified by flash column chromatography, eluting with EtOAc/heptane (18\%, v/v). Solvent removal under reduced pressure yielded 1,2-diphenylpropan-1-ol as a transparent, oily solid (1.99 g).
The crude solid was then dissolved in pentane (10 mL) and allowed to crystallise by slow evaporation over 48 h at room temperature. The resulting crystal bearing oily residue on the surface was transferred into a 15 mL glass vial. Ice-cold pentane (3 mL) was added to submerge the crystal. The vial was gently swirled for 3–5 seconds, after which the supernatant was drawn off. The crystal was rinsed a second time with ice-cold pentane (3 mL) and immediately decanted.  The purified crystal was transferred onto lint-free filter paper and was allowed to dry to constant mass at room temperature, affording the product. Yield: 1687.89 mg (83.6\%).  
\\

\textbf{\textsuperscript{1}H NMR} (400 MHz, \ce{DMSO-d6}): $\delta$ 7.22 – 7.07 (\textit{m}, 10H, Ar\textit{H}), 5.31 (\textit{d}, \textsuperscript{3}$J_\text{H,H'}$ = 4.8 Hz, 1H, (H)CO\textit{H}), 4.64 (\textit{dd}, \textsuperscript{3}$J_\text{H,H'}$ = 4.6 Hz, \textsuperscript{3}$J_\text{H,H'}$ = 6.4 Hz, 1H, (\textit{H})COH), 2.96 (\textit{p}, \textsuperscript{3}$J_\text{H,H'}$ = 6.9 Hz, 1H, (\textit{H})CCH\textsubscript{3}), 1.24 (\textit{d}, \textsuperscript{3}$J_\text{H,H'}$ = 7.0 Hz, 3H, (H)CC\textit{H}\textsubscript{3}).   \\

\textbf{\textsuperscript{13}C\{\textsuperscript{1}H\} NMR} (101 MHz, \ce{DMSO-d6}): $\delta$ 145.41 (arom.), 145.18 (arom.), 128.52 (arom.), 128.23 (arom.), 127.95 (arom.), 126.99 (arom.), 126.90 (arom.), 126.25 (arom.), 77.49 (\textit{C}OH), 47.53 (\textit{C}CH\textsubscript{3}), 16.97 (C\textit{C}H\textsubscript{3})

\begin{figure}[H]
\centering
\includegraphics[width=\textwidth,height=\textheight,keepaspectratio]{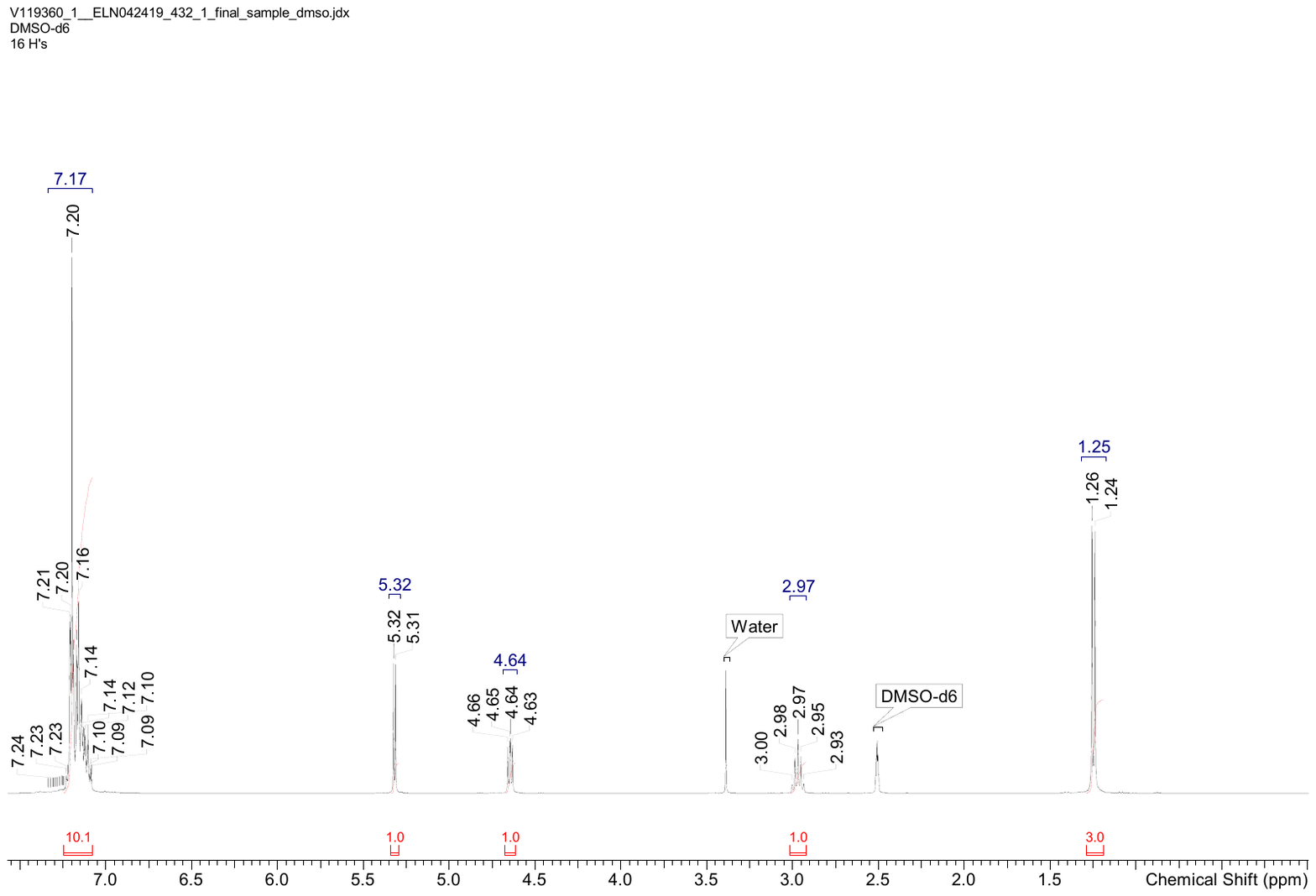}
  \caption{${^1}$H NMR (400 MHz, \ce{DMSO-d6}) of (1\textit{S}, 2\textit{R})-1,2-diphenylpropan-1-ol}
  \label{fgr:HNMR_hydrogenation}
\end{figure}

\begin{figure}[H]
\centering
\includegraphics[width=\textwidth,height=\textheight,keepaspectratio]{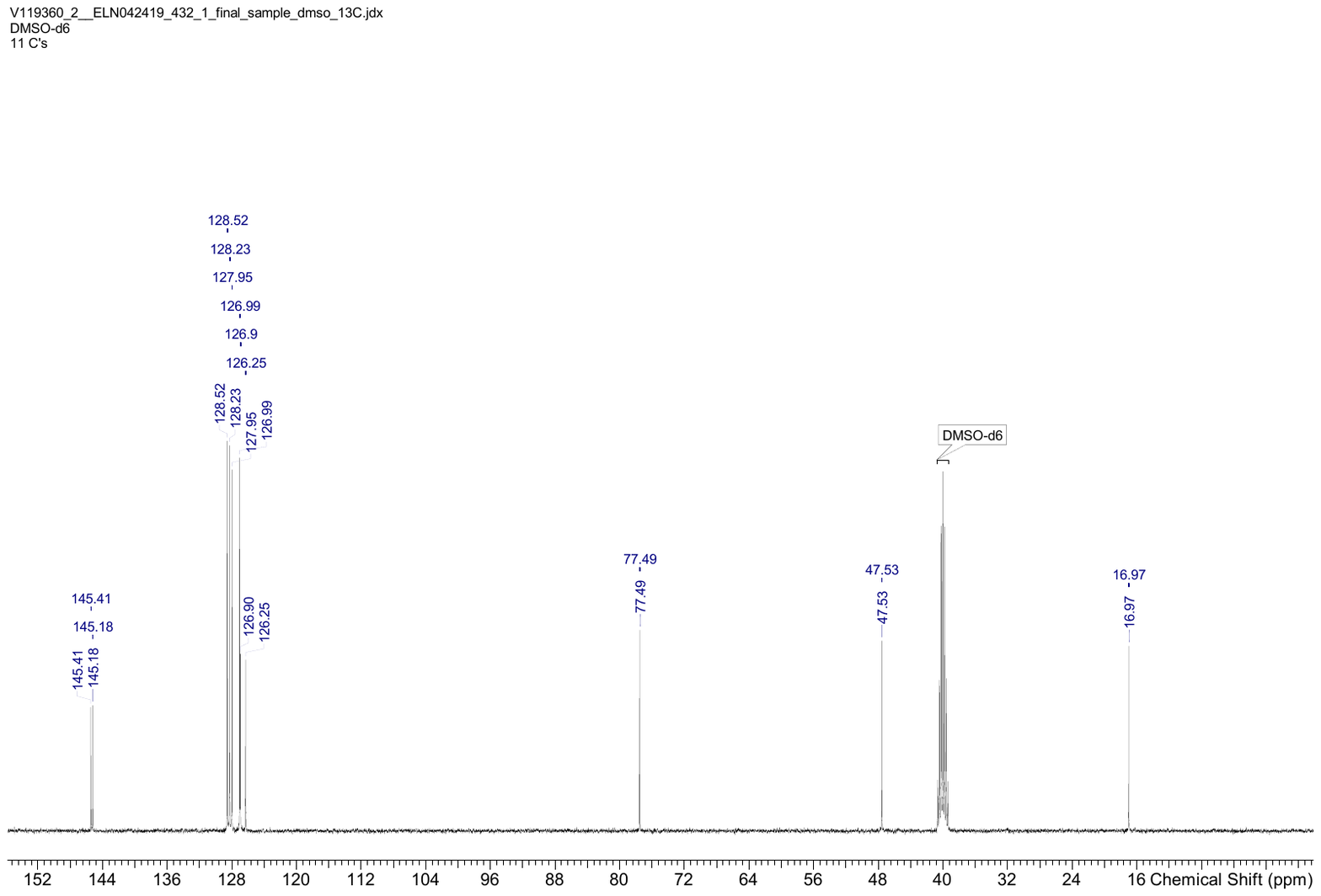}
  \caption{${^{13}}$C\{${^1}$H\} NMR (101 MHz, \ce{DMSO-d6}) of (1\textit{S}, 2\textit{R})-1,2-diphenylpropan-1-ol}
  \label{fgr:CNMR_hydrogenation}
\end{figure}

\newpage

\subsection{Enantiomeric data augmentation}
All solvents and bases in the design space are achiral, and all 32 chiral catalysts are present as enantiomeric pairs. A catalyst and its enantiomer are therefore expected to yield mirror-image product distributions under otherwise identical conditions. Conversion and diastereomeric excess are unchanged, while the enantiomeric excess changes sign. We exploited this symmetry as a data augmentation strategy. For each observed reaction, a reflected counterpart was added to the training data in which the catalyst was replaced by its enantiomer, the sign of ee (\textit{syn}) inverted, and conversion and de (\textit{syn}) retained. Augmented observations were used to fit the surrogate models only. The acquisition function was evaluated over the full design space excluding conditions already run experimentally.

\subsection{Model sensitivity to initial training data}

The first initialisation plate in the HTE reaction optimisation campaign contained only five conditions producing the target \textit{syn} diastereomer. To assess how strongly the round-2 optimisation outcome depended on these observations, we repeated the optimisation with the top \textit{n} conditions removed from the Plate 1 training data, for \textit{n} = 1, 3, and 5, ranking conditions by de (\textit{syn}).\\

In each case the surrogate models were refitted on the reduced dataset and the acquisition function was evaluated over the full design space, excluding conditions already queried. No new experiments were performed: suggested conditions for which experimental data were already available were compared against the best condition remaining in the ablated training set. Because only suggestions already measured in round 2 could be evaluated, the values below are lower bounds on the performance recoverable at each ablation level.\\

With the single most diastereoselective condition removed ($n=1$), the best remaining training observation reached 47.9\% de (\textit{syn}). The model recovered conditions delivering 99.4\% conversion, 75.4\% de (\textit{syn}), and 89.5\% ee (\textit{syn}), improving stereoselectivity. With the top three removed ($n=3$), the best remaining training observation reached 23.5\% de (\textit{syn}), and the model recovered conditions at 64.2\% de (\textit{syn}), again with higher enantioselectivity. Removing all five conditions ($n=5$) that were \textit{syn}-selective left no positive diastereoselectivity examples in the training data, yet the model still recovered conditions reaching 83.4\% de (\textit{syn}) and 93.3\% ee (\textit{syn}). Even with no \textit{syn}-selective examples to learn from, the model identified conditions delivering high \textit{syn} selectivity.

\newpage

\printbibliography